%% file: acl2026_camera_ready.tex
\documentclass[11pt]{article}
\PassOptionsToPackage{table}{xcolor}
\usepackage[]{acl}
\usepackage{times}
\usepackage{latexsym}
\usepackage[T1]{fontenc}
\usepackage[utf8]{inputenc}
\usepackage{inconsolata}
\usepackage{graphicx}
\usepackage{multirow}
\usepackage{tabularx}
\usepackage{booktabs}
\usepackage{longtable}
\usepackage{adjustbox}
\usepackage{caption}
\usepackage[most]{tcolorbox}
\tcbuselibrary{skins, breakable, theorems}
\tcbuselibrary{skins, breakable}
\usepackage{listings}

\usepackage[table,xcdraw]{xcolor}
\newcommand{\tealval}[1]{\textcolor{teal}{#1}}
\newcommand{\redval}[1]{\textcolor{red}{#1}}

 \title{Lost in Translation: Do LVLM Judges Generalize Across Languages?}

\author{Md Tahmid Rahman Laskar\textsuperscript{\textdaggerdbl,}\thanks{\hspace{0.115cm} Contact Emails: \{tahmedge,enamulh,jhuang\}@yorku.ca},  \textbf{Mohammed Saidul Islam\textsuperscript{\textdaggerdbl}}\textbf{,} \\  
 \textbf{Mir Tafseer Nayeem\textsuperscript{\textsection}}\textbf{,} 
\textbf{Md Amran Bhuiyan\textsuperscript{\textdaggerdbl}}\textbf{,} 
\textbf{Mizanur Rahman}\textsuperscript{\textdaggerdbl}\textbf{,} \\
 \textbf{Shafiq Joty\textsuperscript{\textdollar,\textparagraph}}\textbf{, }
 \textbf{Enamul Hoque\textsuperscript{\textdaggerdbl,}}\footnotemark[1]\textbf{, }
 \textbf{Jimmy Xiangji Huang\textsuperscript{\textdaggerdbl,}\footnotemark[1]} \\
 \textsuperscript{\textdaggerdbl}York University, 
 \textsuperscript{\textsection}University of Alberta,\\  
 \textsuperscript{\textdollar}Nanyang Technological University,
 \textsuperscript{\textparagraph}Salesforce AI Research
}

\begin{document}

\maketitle

\begin{abstract}

Automatic evaluators such as reward models play a central role in the alignment and evaluation of large vision–language models (LVLMs). Despite their growing importance, these evaluators are almost exclusively assessed on English-centric benchmarks, leaving open the question of how well these evaluators generalize across languages. To answer this question, we introduce \textbf{MM-JudgeBench}, the first large-scale benchmark for multilingual and multimodal judge model evaluation, which includes over 60K pairwise preference instances spanning 25 typologically diverse languages. MM-JudgeBench integrates two complementary subsets: a general vision–language preference evaluation subset extending VL-RewardBench, and a chart-centric visual–text reasoning subset derived from OpenCQA, enabling systematic analysis of reward models~(i.e., LVLM judges) across diverse settings. \textcolor{black}{We additionally release a multilingual training set derived from MM-RewardBench, disjoint from our evaluation data, to support domain adaptation}. By evaluating 22 LVLMs (15 open-source, 7 proprietary), we uncover substantial cross-lingual performance variance in our proposed benchmark. 
Our analysis further shows that model size and architecture are poor predictors of multilingual robustness, and that even state-of-the-art LVLM judges exhibit inconsistent behavior across languages. Together, these findings expose fundamental limitations of current reward modeling and underscore the necessity of multilingual, multimodal benchmarks for developing reliable automated evaluators.

\end{abstract}

\begin{figure}[t!]
    \centering
\includegraphics[width=\linewidth]{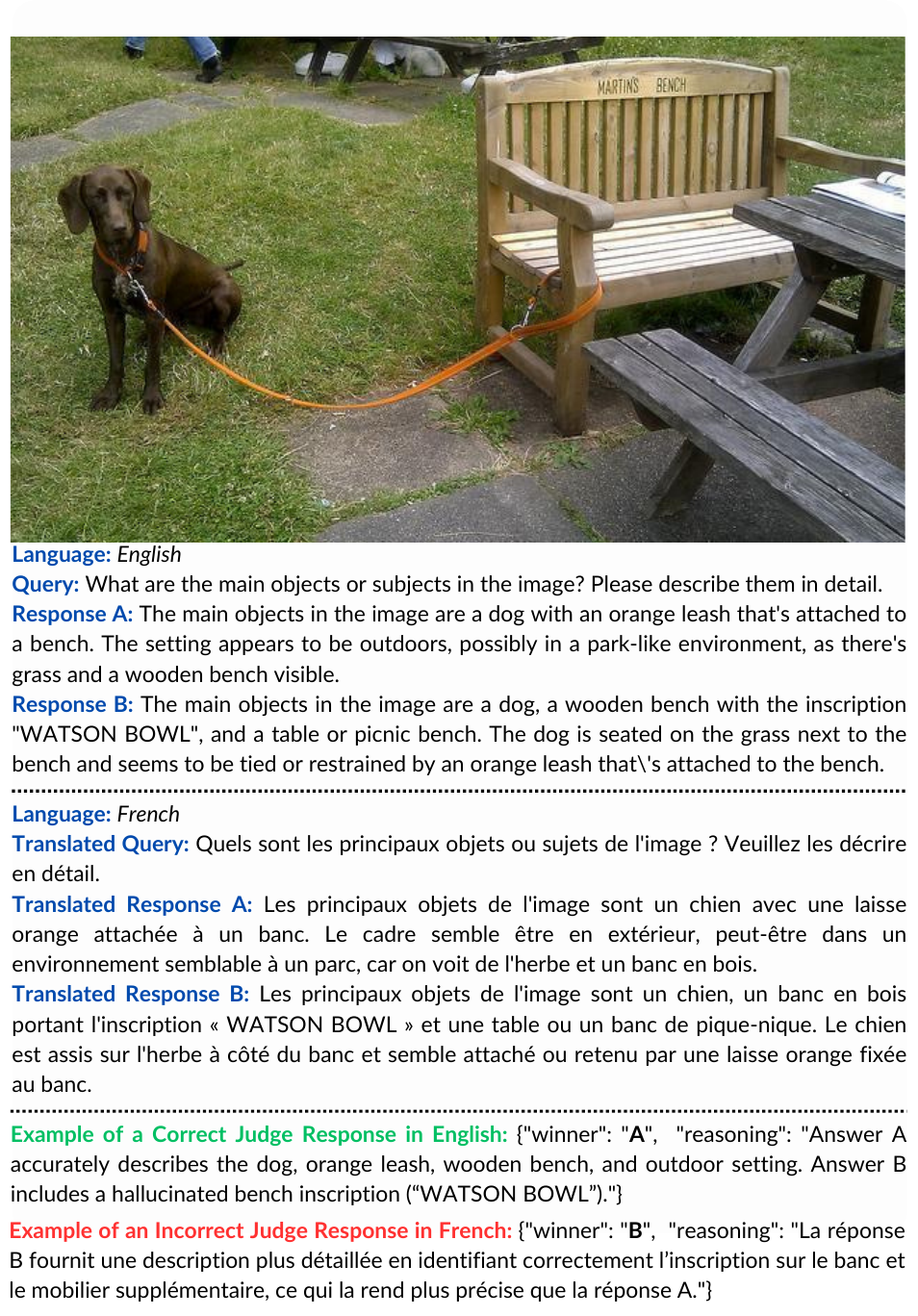}
\caption{\small{Example from VL-RewardBench subset of MM-JudgeBench illustrating multilingual evaluation of LVLM judges for a given image. The question and candidate responses are translated from English to French. The LVLM judge (Gemini-2.5-Flash-Lite) selects the correct response A for English, while incorrectly selects B for French, highlighting the need for multilingual evaluation of LVLM judges.}}
    \label{fig:intro_fig}
\end{figure}
\section{Introduction}
Recent advances in Large Language Models (LLMs) have substantially broadened their multilingual and multimodal capabilities by enabling reasoning over diverse languages and visual modalities \cite{yin2024survey,shohan-etal-2024-xl,qin2025survey}. 
As these systems are deployed at scale, their development and alignment rely critically on \emph{automated evaluators}, most commonly implemented through reward models \cite{ouyang2022training,bai2022constitutional} or LLM/LVLM-as-a-judge frameworks~\cite{zheng2023judging,li2025generation}. 
In practice, automated evaluators now play a central role throughout the LVLM development cycle, from guiding training and alignment to selecting models and benchmarking competing systems.
Despite this central role, the evaluation of reward models or LVLM judges remains overwhelmingly \emph{English-centric}. Vision–language reward benchmarks such as VL-RewardBench \cite{li2025vlrewardbench} and Multimodal RewardBench \cite{yasunaga2025multimodal} focus exclusively on English, while multilingual extensions such as M-RewardBench \cite{gureja2025mrewardbench} are limited to text-only inputs \cite{lambert2025rewardbench}. As a result, no existing benchmark enables a unified study of reward models operating jointly across \emph{languages} and \emph{modalities}, precisely the setting in which these evaluators are expected to be increasingly applied. 

To address this gap, we introduce \textbf{MM-JudgeBench}, the first large-scale benchmark for multilingual and multimodal judge model evaluation. MM-JudgeBench unifies two complementary subsets within a single framework: \textit{(i)} a multilingual extension of VL-RewardBench covering vision–language preference judgments, and \textit{(ii)} a chart-centric visual–text reasoning subset derived from OpenCQA \cite{kantharaj2022opencqa}, which has been widely used to evaluate LVLM judges on structured multimodal inputs \cite{laskar-etal-2025-judging,laskar-etal-2025-deploying}. Together, these subsets span 25 typologically diverse languages and over 60K high-quality preference instances, enabling systematic analysis of LVLM judges across multilingual and multimodal 
settings (see Figure \ref{fig:intro_fig}). To support LVLM judge improvement in multilingual settings, we also release a multilingual training set derived from Multimodal RewardBench 
\cite{yasunaga2025multimodal}.

We construct MM-JudgeBench by translating all benchmark instances using \textit{Gemini-3-Pro}\footnote{\url{https://deepmind.google/models/gemini/pro/}},
selected after empirical validation as a high-quality multilingual translation model. This allows us to isolate cross-lingual evaluation effects while minimizing translation noise. Using this unified benchmark, we conduct a large-scale evaluation of 22 state-of-the-art LVLMs, including leading proprietary models, as well as open-source models spanning model sizes from 1B to 32B parameters.

Our evaluation reveals several key findings. First, while many existing LVLMs report strong average accuracy, this overall score hides meaningful differences across languages, including clear drops in performance for certain languages. For instance, efficiency-optimized model variants often suffer severe multilingual performance collapse despite strong English performance. Second, among open models, 
Qwen3-VL exhibits 
the most consistent multilingual behavior, outperforming many larger alternatives. Third, beyond accuracy, we uncover pronounced 
biases and instruction-following failures, demonstrating that correctness alone is insufficient to guarantee reliable reward-based evaluation. Finally, we 
show that reasoning-augmented judging 
and domain-adaptive fine-tuning on multilingual reward data 
offer noticeable performance gain. 
Our main contributions are listed as follows:
 \vspace{-1mm}
\begin{itemize}
\item \textbf{MM-JudgeBench}, the first large-scale benchmark for \emph{multilingual and multimodal} evaluation of LVLM judges, covering VL-RewardBench 
and OpenCQA across 25 typologically diverse languages and over 60K preference instances. 
\vspace{-2mm}
\textcolor{black}{
\item A \textbf{large-scale empirical evaluation of 22 state-of-the-art LVLMs}, spanning both proprietary and open models across a wide range of model sizes and architectures, revealing cross-lingual performance variations that are invisible under English-only evaluation. In addition, we reveal the scaling behavior of LVLM judges in multilingual settings, alongside various biases. 
\vspace{-2mm}
\item 
In addition, we release a \textit{multilingual training set covering 100k preference instances} derived from MM-RewardBench to support \textbf{domain adaptive fine-tuning} as an improvement strategy for cost-efficient open models.}
\end{itemize}
 \vspace{-1mm}
To support reproducibility and further research, we publicly release MM-JudgeBench along with the evaluation code 
at \url{https://github.com/tahmedge/mm-judgebench}.

\begin{figure*}[t!]
    \centering
    \includegraphics[width=\linewidth]{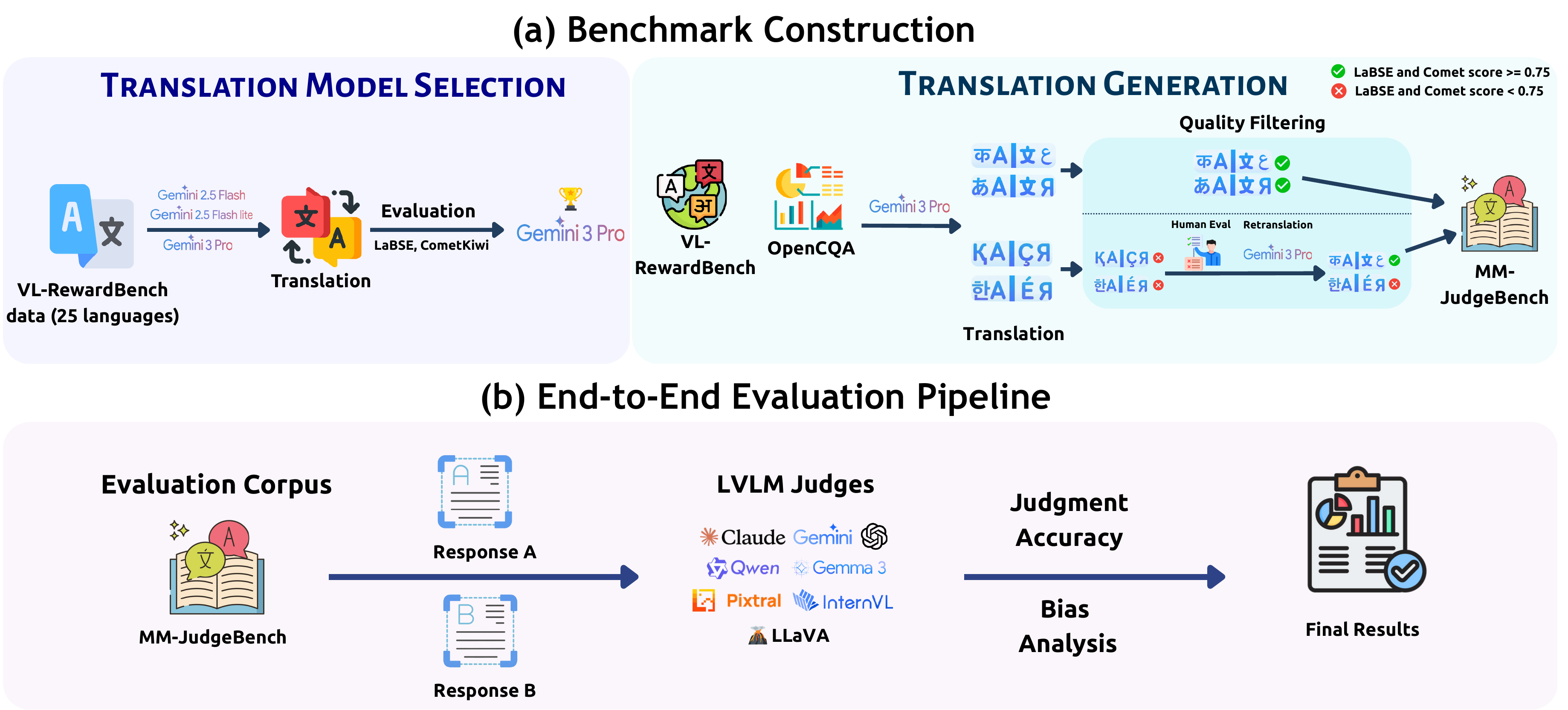}
    \caption{{An overview of our methodology: \textit{(a) Benchmark construction} step contains two stages, i.e., Translation model selection, and Translation data generation (from VL-Reward Bench and OpenCQA data); and \textit{(b) End-to-End evaluation pipeline}.}}
    \vspace{-2mm}
    \label{fig:method}
\end{figure*}

\vspace{-1mm}
\section{Related Work}
\vspace{-1mm}


Reward models are a cornerstone of modern LLM development \cite{laskar-etal-2023-systematic,jahan-etal-2023-evaluation}, particularly for aligning models with human preferences  \cite{christiano2017deep, ouyang2022training}. As a result, the quality of the reward model directly influences not only the downstream performance of aligned LLMs but also their safety and reliability. This dependency has elevated the evaluation of reward models as a critical research problem, motivating the development of dedicated evaluation benchmarks. 

The first systematic benchmark proposed for this purpose was RewardBench \cite{lambert2025rewardbench}. However, this benchmark was restricted to text-based modality on English-centric data. Consequently, the rapid advancement in multilingual and multimodal LLMs has underscored the need for benchmarks that can robustly assess model capabilities across diverse languages and input modalities \cite{yin2024survey,qin2025survey}. To this end, benchmarks like VL-RewardBench \cite{li2025vlrewardbench} and M-RewardBench \cite{gureja2025mrewardbench} attempted to mitigate this gap by utilizing multimodal and multilingual data, respectively.

However, the above-mentioned benchmarks are also limited in scope.
For instance, while VL-RewardBench provides a challenging set of examples for evaluating vision-language reward models across complex reasoning tasks, it is restricted to only English-centric data. On the other hand, while M-RewardBench highlights the need for linguistically diverse benchmarks by extending RewardBench to a multilingual setting, it is restricted to text-based modality. To our best knowledge, there are no multimodal reward benchmarks that are also multilingual.
Our work addresses this gap 
by directly extending VL-RewardBench by translating it  
across 25 languages, providing a comprehensive multilingual benchmark for the evaluation of multimodal reward models.

Beyond reward modeling, the closely related LVLM-as-a-judge paradigm has emerged, where capable LVLMs serve as evaluators for responses generated by other vision-language systems \cite{chen2024mllm}. Despite growing adoption, existing studies evaluating LVLM judges are also limited to the general multimodal task evaluation in English \cite{lee2024prometheus,xiong2025llava}. We address this gap by utilizing the OpenCQA dataset \cite{kantharaj2022opencqa}, which is an open-domain chart question answering dataset and used by \citet{laskar-etal-2025-judging,laskar-etal-2025-deploying} for the evaluation of LVLMs-as-the-judge.
We extend our contribution to this paradigm by also translating OpenCQA across the same 25 languages, enabling more comprehensive cross-lingual assessment of LVLM judges, as recent work 
underscores the need for chart-centered evaluations of LVLMs~\cite{11298756,mahbub-etal-2025-charts,laskar-etal-2025-judging,laskar-etal-2025-deploying}.

\textcolor{black}{Beyond proprietary judges, open-source models such as LLaVA-Critic \cite{xiong2025llava} and Prometheus-Vision \cite{lee2024prometheus} have been trained specifically for vision–language evaluation. However, these models are developed and assessed exclusively in English. 
To support the development of multilingual multimodal judge models, we construct the multilingual version of MM-RewardBench \cite{yasunaga2025multimodal} by extending it into 24 languages.}



%


\vspace{-1mm}
\section{Methodology}
\vspace{-1mm}
Our methodology for creating and evaluating the multilingual multimodal reward model benchmark involved \emph{three} main stages: translation model selection, dataset construction, and model evaluation. Figure \ref{fig:method} demonstrates an overview of our methodology, which we also describe below (furthermore, section \ref{domain_adaptive_fine_tuning} discusses our domain adaptive fine-tuning approach).

 \vspace{-1mm}
\subsection{Translation Model Selection}
 \vspace{-1mm}
To create a comprehensive multilingual benchmark, we follow the work of \citet{gureja2025mrewardbench} and selected the 23 languages they used in M-RewardBench. However, since none of their selected languages are considered low-resource, we added two languages: \textit{Bengali} and the low-resource \textit{Kazakh} to ensure a more robust evaluation of LVLMs. 
The selected 25 languages cover a diverse range of language families and scripts: \textit{Arabic (ar), Bengali (bn), Chinese (zh), Czech (cs), Dutch (nl), English (en), French (fr), German (de), Greek (el), Hebrew (he), Hindi (hi), Indonesian (id), Italian (it), Japanese (ja), Kazakh (kk), Korean (ko), Persian (fa), Polish (pl), Portuguese (pt), Romanian (ro), Russian (ru), Spanish (es), Turkish (tr), Ukrainian (uk),} and \textit{Vietnamese (vi)}.

For the translation process, we utilized Gemini-3-Pro. 
We select this model for translation due to the impressive performance of the Pro version of Gemini 
in WMT25 \cite{kocmi2025findings}.  

To validate our choice of Gemini-3-Pro as the translation model, we conducted a comparative analysis of translation quality across 25 languages in VL-RewardBench against Gemini-2.5-Pro alongside two other cheaper Gemini variants: Gemini-2.5-Flash and Gemini-2.5-Flash-Lite. First, the query and the pairwise responses were translated. Then, the quality of the translations was assessed using two standard translation quality estimation metrics: Language-agnostic BERT Sentence Embedding (LaBSE) \cite{feng-etal-2022-language} and CometKiwi \cite{rei2022cometkiwi}. As shown in Table \ref{tab:translation_quality}, the Pro versions (Gemini-3-Pro and Gemini-2.5-Pro) achieved the highest scores on both metrics, confirming their suitability for translation (see Appendix \ref{trans_quality_eval_opencqa} for the translation quality of OpenCQA). While both Gemini-3-Pro and Gemini-2.5-Pro perform identically, we selected Gemini-3-Pro since it is the latest version of the Gemini-Pro series. 
This helps ensure long-term reproducibility, as older closed models are often deprecated.

 \vspace{-1mm}
\subsection{Data Construction}
\vspace{-1mm}
\noindent \textbf{Translation process.} For data translation, in each dataset, we create a prompt for Gemini-3-Pro to translate the query (if available) and the corresponding answers generated by two different models 
for each image from English to 24 other languages. For VL-RewardBench and MM-RewardBench, we translate the answers that are publicly available in the dataset. For OpenCQA, we use the answers generated using Gemini-1-Pro and Claude-3-Haiku by \citet{islam2024large} and then translate them. For all benchmarks, we translate the queries provided in each dataset. We denote the multilingual (M) versions as M-VL-RewardBench, M-OpenCQA, and M-MM-RewardBench. 

\noindent \textbf{Prompt design and API efficiency.} Given the per-day rate limit of API calls\footnote{\url{https://ai.google.dev/gemini-api/docs/rate-limits}} for powerful closed models, we ask the translator LLM (e.g., Gemini-3-Pro) to translate the query (where available) and the answers from English to 24 other languages within a single prompt (see Appendix \ref{translation_details} for details). The translator LLM was required to generate the response in an Array of JSON objects format 
with the following keys: \textbf{(i)} Translated Query, \textbf{(ii)} Translated Answer A, \textbf{(iii)} Translated Answer B, \textbf{(iv)} Language. Then, we write a parsing script to extract the translated data for each language from the JSON-formatted output. This enables the reduction of API calls by 24 times. 

\noindent \textbf{Quality filtering and final datasets.} To ensure high quality of the translated datasets, 
we have also added a quality filtering step. \textcolor{black}{Inspired by prior threshold-based filtering work \cite{schwenk2018filtering,batheja2023little}, 
samples having LaBSE and Comet scores below 0.75 are first manually inspected by a human via back translation.} If the human reviewer finds that the translation is bad, we re-translate that sample by tuning the decoding parameters. After re-translation, if the score is still below 0.75, we remove that sample. Our final datasets contain 31K samples from VL-RewardBench, 30K samples from OpenCQA, and over 100K samples from MM-RewardBench.

\begin{table}[t]
\footnotesize
\centering
\begin{tabular}{lcc}
\toprule
\rowcolor[HTML]{ECEFF1}
\textbf{Model} & \textbf{LaBSE} & \textbf{CometKiwi} \\
\midrule
\rowcolor[HTML]{FAFAFA}
Gemini-3-Pro & 0.91 & 0.85 \\
\rowcolor[HTML]{FAFAFA}
Gemini-2.5-Pro & 0.91 & 0.85 \\
\rowcolor[HTML]{FAFAFA}
Gemini-2.5-Flash & 0.88 & 0.83 \\
\rowcolor[HTML]{FAFAFA}
Gemini-2.5-Flash-Lite & 0.79 & 0.74 \\
\bottomrule
\end{tabular}
\vspace{-2mm}
\caption{\small{Translation quality comparison of Gemini models in VL-RewardBench.}}
\label{tab:translation_quality}
\end{table}

 \vspace{-1mm}
\subsection{LVLM Evaluation}
 \vspace{-1mm}
We select a wide range of both proprietary and open-source vision-language models to evaluate in our proposed MM-JudgeBench benchmark. The models include:  GPT-5\footnote{\url{https://openai.com/gpt-5/}} and its variants (Mini and Nano), Gemini-2.5-Flash and Gemini-2.5-Flash-Lite \cite{comanici2025gemini}, Claude-4.5-Haiku\footnote{\url{https://www.anthropic.com/claude/haiku}}, Grok-4.1-Fast\footnote{\url{https://x.ai/news/grok-4-1-fast}}, 
 Gemma-3 (4B, 12B, 27B) \cite{team2025gemma},
 InternVL-3.5 \cite{wang2025internvl3} (1B, 2B, 4B, 8B, 14B),
 Qwen3-VL \cite{Qwen3-VL} (2B, 30B-A3B, 4B, 8B, 32B), and 
 Pixtral-12B \cite{agrawal2024pixtral}. We also select the  LLaVA-Critic-7B \cite{xiong2025llava} model, which is an LVLM trained for reward modeling. 

For each model and language, we provide the query and two candidate answers to the LVLM judge and prompt it to select the better answer. In addition, the LVLM is also asked to provide a rationale behind their selection since adding rationale helps improve the performance of reward models or judges \cite{xiong2025llava}.
\definecolor{attachedColor5}{HTML}{ECEFF1}
\definecolor{attachedColor}{HTML}{e0efff}
\definecolor{attachedColor2}{HTML}{f3f3f3}
\definecolor{attachedColor3}{HTML}{FFE5CC}
\definecolor{attachedColor4}{HTML}{FFCCCC}
\begin{tcolorbox}[
boxrule=0.25pt,   
  colback=attachedColor2,    
  colframe=black,           
  colbacktitle=attachedColor, 
  coltitle=black,           
  title={{LVLM Judge Prompt}},
  fonttitle=\bfseries,      
  fontupper=\small          
]

You are a strict and fair judge for vision-language tasks. You will be shown an image and a user question, plus two candidate answers A and B. Decide which answer is better based on the following criteria. \\

- Correctness with respect to the image and question.\\
- Completeness and level of detail.\\
- Relevance and clarity (no unnecessary verbosity).\\
- Avoiding hallucinations or unsupported claims.\\

Return ONLY a JSON object with this schema:\\
\{``winner'': ``A | B'', ``reasoning'': ``brief reason''\}

\end{tcolorbox}

For evaluation, we measure the pairwise accuracy, which is the percentage of times the LVLM correctly identified the preferred response in a given pair. Additionally, we investigate the presence of biases like position and length bias. \textit{Position bias} refers to the tendency of a model to prefer answers based on their position in the input (e.g., always choosing the first or second response). To measure this, we present each pair of answers to the models in both the original and a reversed order. The position bias is then calculated as the overall difference in judgment accuracy between the two orderings. \textit{Length bias} is measured by evaluating whether the model prefers an answer that is longer than the other answer but incorrect.






\input{vlrewardbench_results_table.tex}
 \vspace{-1mm}
\section{Results and Discussion}
 \vspace{-1mm}
We conduct a series of experiments to evaluate various LVLM judges in MM-JudgeBench. 

\noindent \textbf{Evaluation Settings:} The evaluation considers assessing the pairwise accuracy, focusing on factual correctness, completeness, and relevance. {To ensure robustness, we generate the judgment by providing the candidate answers in both the original and the reversed order independently, and then report the average accuracy.}
In addition, we evaluate biases like positional bias and length bias. As the LVLMs generate the winner alongside the reasoning, we parse the LVLM-judge predicted winner from their corresponding JSON-formatted responses using a parsing script \cite{laskar-etal-2024-systematic}. If the parsing script cannot properly parse the judgment from the response, we consider the LVLM-generated answer wrong. All the open models were run using vLLM \cite{kwon2023efficient} on an A100 GPU, with the decoding parameters being primarily set to the default values in HuggingFace \cite{wolf-etal-2020-transformers} for the respective open-source models. For the closed models, decoding parameters are also set to the default values as provided by the respective model providers. For all supporting models, the temperature value is set to 0. Below, we demonstrate our findings.


 \vspace{-1mm}
\subsection{Main Results on M-VL-RewardBench
}\vspace{-1mm}
Table~\ref{tab:vlrewardbench_results} reports the accuracy of vision–language models evaluated on the M-VL-RewardBench subset of MM-JudgeBench across 25 languages. In addition, we report the per-model average (Avg) and variance (Var) across languages. 
Below, we summarize our key findings.


\noindent \textbf{Closed models dominate, but efficient variants are uneven.} GPT-5 achieves the highest overall accuracy (81.3\%) with the lowest variance (0.2), remaining stable from the high-resource English (81.7\%) to the low-resource Kazakh (80.6\%). Cost-effective closed models such as GPT-5-Mini, GPT-5-Nano, Gemini-2.5-Flash, and Grok-4.1-Fast retain reasonable accuracy (71–78\%) with variance mostly below 1 (except GPT-5-Nano). However, some cost-efficient variants perform quite poorly: Gemini-2.5-Flash-Lite collapses to 40.8\% (variance 2.6), and Claude-4.5-Haiku drops to 56.4\% (variance 2.1). 


\noindent \textbf{Qwen3-VL is the strongest open family and scales consistently.} Qwen3-VL-32B achieves the best performance among open models (68.8\% on M-VL-RewardBench), with accuracy improving monotonically from 54.3\% at 2B to 68.8\% at 32B. Even the smaller Qwen3-VL-4B (61.9\%) surpasses all InternVL-3.5 and Gemma-3 variants as well as Pixtral-12B, and several Qwen3 variants outperform the weaker closed models (Claude-4.5-Haiku, Gemini-2.5-Flash-Lite). Surprisingly, the reward-specialized LLaVA-Critic-7B scores below 50\%, suggesting poor generalization. 
\noindent \textbf{Closed-source versus open-source gap.}
Our evaluation reveals a clear performance gap between the strongest closed and open models. For instance, the most optimized GPT-5 variant, GPT-5-Nano, outperforms the best open model, Qwen3-VL-32B, by approximately 4.5 points on average. However, many open models (e.g., Qwen3) substantially outperform some closed models (e.g., Claude-4.5-Haiku and Gemini-2.5-Flash-Lite). This indicates that architectural design and training strategies could play a critical role (see Appendix \ref{performance_script_groups} for further analysis). More broadly, upstream tokenization likely also contributes, since compression-oriented metrics alone miss systematic cross-lingual fragmentation ~\cite{alqahtani-etal-2026-stop,nayeem2025fertility}. 

\noindent \textbf{Performance variation across languages.}
A key finding from our evaluation on the VL-RewardBench subset is the inclusion of the low-resource Kazakh language, in which most LVLMs achieve the poorest performance, as demonstrated in color red in Table \ref{tab:vlrewardbench_results}. This suggests that current training strategies may not adequately capture the linguistic diversity required for robust multilingual reward model performance. We also observe that English tends to achieve the best accuracy in comparison to other languages for most models.


\input{opencqa_results_table.tex}
\vspace{-1mm}
\subsection{Main Results on M-OpenCQA}
\vspace{-1mm}
We next evaluate LVLM judges on the M-OpenCQA subset of MM-JudgeBench that focuses on chart-centric visual–text reasoning. Unlike M-VL-RewardBench, M-OpenCQA does not provide gold preference labels. To enable systematic evaluation, we select GPT-5, which is the strongest model on M-VL-RewardBench, as a high-quality reference judge to annotate pairwise preferences\footnote{We evaluate GPT-5 judgments on M-OpenCQA using 3 human evaluators having expertise in NLP, Data Science, and Computer Vision. We randomly collect 300 samples from OpenCQA and find that 93.5\% of the time, at least 2 annotators agree with GPT-5 judgment.}. Consequently, we restrict this analysis to open-source LVLMs, examining how well they can replicate the judgments of a powerful proprietary model. Results are reported in Table \ref{tab:opencqa_results}.

\noindent \textbf{Overall performance trends.}
Across model families, performance on M-OpenCQA in terms of both accuracy and variance is usually better than on M-VL-RewardBench. 
This may reflect the increased difficulty in M-VL-RewardBench, which consists of diverse multimodal tasks. 

\noindent \textbf{Qwen-3-VL again leads among open models.}
Among open models, 
Qwen3-VL-32B again achieves the best result (accuracy 67.4\%, variance 1.4), with all Qwen3 variants above 2B exceeding 60\% while also scaling consistently from 2B to 32B. InternVL-14B is competitive on average (66.3\%) but less stable (variance 2.1). Other open families lag behind, with all Gemma-3 models, Pixtral-12B, as well as the specialized LLaVA-Critic-7B, achieving below 60\%. In summary, these trends mirror M-VL-RewardBench, confirming Qwen3's robustness across datasets.



\noindent \textbf{Language-specific patterns persist.}
Similar to M-VL-RewardBench, English consistently achieves the highest accuracy across most models (13 out of 15 times). Meanwhile, low-resource languages like Kazakh continue to present challenges, with most models achieving their lowest scores on this language (10 out of 15 times). However, the variance in per-language performance is generally lower on M-OpenCQA compared to M-VL-RewardBench, suggesting that chart reasoning is less sensitive to language-specific factors.


\begin{table}
\tiny
\setlength{\tabcolsep}{4pt}
\centering
\begin{tabular}{lccc}
\toprule
\rowcolor[HTML]{ECEFF1}\textbf{Model} & \textbf{M-OpenCQA} & \textbf{M-VL-RewardBench} & \textbf{Average} \\
\midrule
Qwen3-VL-32B & 1.28 | \tealval{0.43} & 5.58 | \tealval{3.91} & 3.43 | \tealval{2.18} \\
Qwen3-VL-8B & 1.98 | 0.84 & \tealval{2.49} | 7.40 & \tealval{2.24} | 4.12 \\
Qwen3-VL-4B & \tealval{0.91} | 0.59 & 36.36 | 43.97 & 18.64 | 22.28 \\
Qwen3-VL-30B-A3B & 3.42 | 0.43 & 9.00 | 7.00 & 6.21 | 3.72 \\
Qwen3-VL-2B & 2.90 | 1.57 & 18.12 | {52.94} & 10.51 | {27.26} \\
InternVL-3.5-14B & 2.34 | 1.02 & 5.35 | 14.36 & 3.84 | 7.69 \\
InternVL-3.5-8B & 1.76 | 1.43 & 22.92 | 21.10 & 12.34 | 11.26 \\
InternVL-3.5-4B & 2.17 | 1.40 & 38.45 | 27.96 & 20.31 | 14.68 \\
InternVL-3.5-2B & 5.87 | \redval{3.26} & \redval{55.90} | 25.95 & \redval{30.88} | 14.60 \\
InternVL-3.5-1B & \redval{6.20} | 2.44 & 30.60 | {136.10} & 18.40 | {69.27} \\
Gemma-3-27B & 1.47 | 0.47 & 8.24 | 6.53 & 4.86 | 3.50 \\
Gemma-3-12B & {0.92} | 0.57 & 8.07 | 8.79 & 4.50 | 4.68 \\
Gemma-3-4B & 1.19 | 0.44 & 12.94 | 18.13 & 7.06 | 9.28 \\
LLaVA-Critic-7B & 1.98 | 2.68 & 15.82 | \redval{180.05} & 8.90 | \redval{91.36} \\
Pixtral-12B & 3.50 | 1.71 & 40.56 | 11.93 & 22.03 | 6.82 \\
\bottomrule
\end{tabular}
\vspace{-2mm}
\caption{\small{Avg. Position Bias and Variance (separated by `|', \textit{left} part denotes position bias and \textit{right} part denotes position bias variance) for the open-source LVLMs in the M-VL-RewardBench. Lower values indicate better with \textcolor{teal}{Green} indicates the best and \textcolor{red}{Red} indicates the worst, per column.}}
\label{tab:pos_bias_variance}
\end{table}

 \vspace{-1mm}
\subsection{Bias Analysis}  
\begin{figure}[t!]
    \centering
    \includegraphics[width=\linewidth, height=4cm]{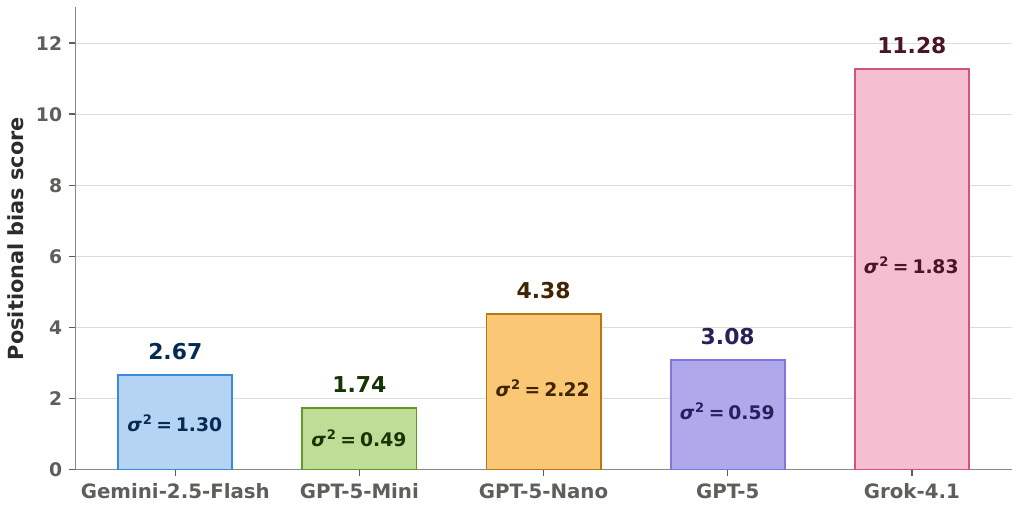}
     \vspace{-7mm}
    \caption{\small{Position Bias in multilingual M-VL-RewardBench for closed models. Lower values indicate better.}}
    \label{fig:pos_bias}
\end{figure}

\vspace{-1mm}
Table \ref{tab:pos_bias_variance} summarizes the average positional bias for open-source LVLMs on both M-VL-RewardBench and M-OpenCQA. In terms of the average across both datasets, we find that the Qwen3-VL-8B is the least prone to positional bias (only 2.24\%).  Moreover, Qwen3-VL-32B shows the lowest bias variance (2.18\%), demonstrating 
judgment capability with less bias across languages. 
In summary, we observe that M-OpenCQA usually has lower position bias and variance in comparison to M-VL-RewardBench. 
Figure \ref{fig:pos_bias} further shows positional bias trends for top-performing closed models (above 70\% accuracy) on M-VL-RewardBench. We observe that the 
GPT-5-Mini exhibits the lowest positional bias, whereas Grok-4.1-Fast shows very high positional bias. Other models (GPT-5 and Gemini-2.5-Flash) also demonstrate low position bias and variance. 
Taken together, these results reveal that robustness to positional coherence is highly model- and dataset-dependent, with only a small subset of LVLMs demonstrating bias-resilient judging behavior across multilingual and multimodal settings. These highlight that only accuracy is insufficient to guarantee reliable reward modeling. In Appendix \ref{cross_lingual_amplification_pos_bias}, we further analyze the cross-lingual amplification of the positional bias. In Appendix \ref{length-bias-appendix}, we also report the length bias. 

 \vspace{-2mm}
\subsection{Impact of Reasoning}
 \vspace{-1mm}
For each LVLM judge, we follow prior work \cite{laskar-etal-2025-judging} to generate the preferred answer alongside the reasoning behind choosing the particular answer in JSON format (Appendix \ref{output_format_instruction_appendix} also evaluates the capabilities of different models in generating the output in the required format).  We further investigate the performance by comparing the direct generation of the judgment against the generation of the judgment alongside the reasoning. We evaluate the Qwen3-VL-2B and Qwen3-VL-8B models with and without reasoning to find that the overall accuracy is degraded by 2.2\% and 4\%, respectively, if no reasoning is used when LVLMs generate the judgment. This validates the utilization of reasoning in judgment.  

In Table \ref{tab:reasoning_labse_scores_compact}, we further evaluate the reasoning quality via LaBSE scores on M-OpenCQA and find that Qwen3-VL consistently outperforms comparable InternVL models (71.64 vs. 64.59 at 8B) and shows consistent scaling (70.70 at 4B to 71.66 at 32B) with diminishing returns beyond 8B. InternVL exhibits noticeable gain from 4B (47.49) to 8B (64.59), but a narrow gain to 14B. 

\begin{table}[t!]
\footnotesize
\centering
\rowcolors{1}{white}{white}
\setlength{\tabcolsep}{4pt}
\begin{tabular}{lccc}
\toprule
\rowcolor[HTML]{ECEFF1}\textbf{Model Family} & \textbf{4B} & \textbf{8B} & \textbf{32B/14B} \\
\midrule
\rowcolor[HTML]{FAFAFA} Qwen3-VL-Instruct   & 70.70 & 71.64 & 71.66 \\
 \rowcolor[HTML]{FAFAFA} Intern-VL-Instruct  & 47.49 & 64.59 & 64.95 \\
\bottomrule
\end{tabular}
\vspace{-2mm}
\caption{\small{Reasoning quality in M-OpenCQA in terms of LaBSE similarity scores for the top-performing open-source models (Qwen3 and InternVL-3.5) across different sizes. Here, 32B denotes Qwen3 while 14B denotes InternVL.}}
\label{tab:reasoning_labse_scores_compact}
\end{table}


 \vspace{-1mm}
\subsection{Domain Adaptation via Fine-Tuning}  
\label{domain_adaptive_fine_tuning}
 \vspace{-1mm}
We fine-tune the Qwen3-VL-8B-Instruct model in our training subset: M-MM-RewardBench. Since M-MM-RewardBench does not have the reference reasoning for the corresponding judgment, we apply supervised fine-tuning only on the judgment labels and train the model to directly predict the judgment without any reasoning. We ran a total of 2 epochs, with the learning rate being set to $2e-5$. On M-VL-RewardBench, the fine-tuned model improves by +14\% over direct prompting and +10\% over rationale-augmented prompting (both in zero-shot). These substantial improvements demonstrate the effectiveness of our training subset in improving the judging performance across other datasets, opening up the possibility of using weak supervision techniques to further improve the performance \cite{laskar-etal-2020-wsl, laskar2022domain}. 

 \vspace{-1mm}
\subsection{Task-Level Performance Analysis}
 \vspace{-1mm}
The M-VL-RewardBench subset contains three task categories from the original VL-RewardBench: (i) General Multimodal Instructions, (ii) Hallucination-Oriented Queries, and (iii) Mathematical Reasoning. Below, we conduct a task-level performance analysis. 

Based on the average across all models, we find that hallucination-oriented queries have an average accuracy of 73.42, followed by mathematical reasoning (68.51) and general multimodal instructions (49.71).
These results indicate that structured evaluation scenarios (hallucination detection, mathematical verification) are currently better suited to LVLM-as-a-judge evaluation, while open-ended multimodal instruction tasks remain challenging. 
Table \ref{tab:task_level_average} further reports task-level accuracy for the best-performing closed (GPT-5) and open (Qwen3-VL) model series. 
We observe that even the best-performing GPT-5 model achieves only about 60\% accuracy in the general multimodal instruction following task. We further demonstrate the English vs. non-English performance gap across tasks in Appendix \ref{task_level_performance_gap}.

\begin{table}[t!]
\centering
\small

\rowcolors{1}{white}{white}
\setlength{\tabcolsep}{2pt}
\begin{tabular}{lccc}
\toprule
\rowcolor[HTML]{ECEFF1}
\textbf{Model} & \textbf{Hallucination} & \textbf{Mathematical} & \textbf{General} \\
\midrule
\rowcolor[HTML]{FAFAFA} Qwen3-VL-32B & 73.42 & 68.51 & 49.71 \\
\rowcolor[HTML]{FAFAFA} Qwen3-VL-8B  & 65.14 & 64.39 & 47.51 \\
\rowcolor[HTML]{FAFAFA} Qwen3-VL-4B  & 64.66 & 63.06 & 48.75 \\
\rowcolor[HTML]{FAFAFA} Qwen3-VL-2B  & 55.59 & 58.01 & 44.15 \\
 \rowcolor[HTML]{FAFAFA} GPT-5        & 83.55 & 87.68 & 60.94 \\
\rowcolor[HTML]{FAFAFA} GPT-5-Mini   & 81.26 & 81.83 & 58.31 \\
\rowcolor[HTML]{FAFAFA} GPT-5-Nano   & 76.61 & 74.45 & 56.56 \\
\bottomrule
\end{tabular}
\vspace{-2mm}
\caption{\small{Per Task Accuracy by Model in M-VL-RewardBench.}}
\label{tab:task_level_average}
\end{table}

\begin{figure*}[t!]
    \centering
\includegraphics[width=\linewidth, height=5.75cm]{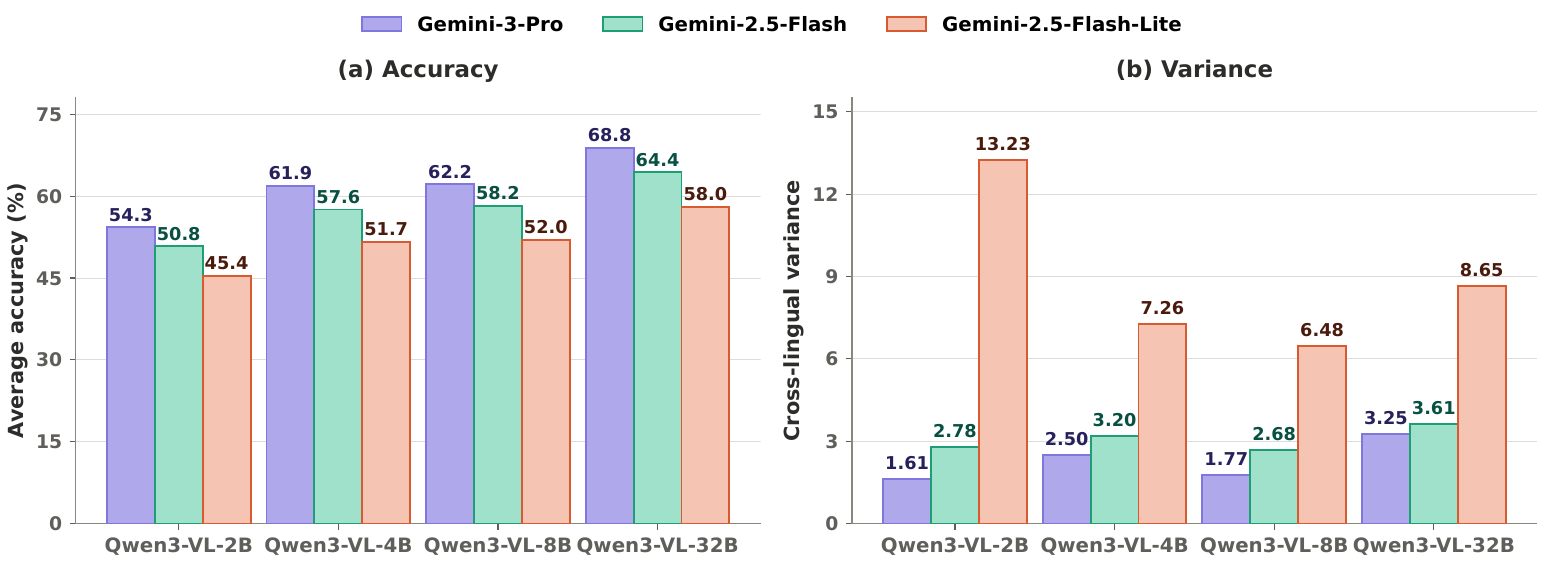}
     \vspace{-7mm}
    \caption{{Translation sensitivity analysis on M-VL-RewardBench with Gemini models as the translator and Qwen3-VL as the judge. Higher is better for Accuracy, while Lower is better for Variance.}}
    \label{fig:translation-sensitivity}
\end{figure*}

 \vspace{-1mm}
\subsection{Translation Sensitivity Analysis}
 \vspace{-1mm}
To verify that our findings are not artifacts of a specific translation system, we repeat the Qwen3-VL evaluation on M-VL-RewardBench using two weaker translators (Gemini-2.5-Flash and Gemini-2.5-Flash-Lite) in addition to our default Gemini-3-Pro. Figure~\ref{fig:translation-sensitivity} shows that weaker translations monotonically reduce average accuracy (e.g., Qwen3-VL-32B: 68.8 $\rightarrow$ 64.4 $\rightarrow$ 58.1) and inflate cross-lingual variance sharply, with Qwen3-VL-2B variance jumping from 1.6 to 13.2 under the weakest translator. Crucially, {model rankings remain preserved across all three translation systems}, indicating that the cross-lingual trends reported in Tables~\ref{tab:vlrewardbench_results} and \ref{tab:opencqa_results} reflect genuine judge behavior rather than translation artifacts.


  \begin{table*}[t]
  \centering                                                                                                                                               
  \tiny           
  \setlength{\tabcolsep}{2pt}                                                                                                                              
  \renewcommand{\arraystretch}{1}
  \begin{tabular}{lccccccccccccccccccccccccccc}                                                                                                            
  \toprule                                                                                                                                                 
  \rowcolor[HTML]{ECEFF1}
  \textbf{Model} & \textbf{Avg} & \textbf{Var} & \textit{ar} & \textit{bn} & \textit{zh} & \textit{cs} & \textit{nl} & \textit{en} & \textit{fr} &         
  \textit{de} & \textit{el} & \textit{he} & \textit{hi} & \textit{id} & \textit{it} & \textit{ja} & \textit{kk} & \textit{ko} & \textit{fa} & \textit{pl} &
   \textit{pt} & \textit{ro} & \textit{ru} & \textit{es} & \textit{tr} & \textit{uk} & \textit{vi} \\                                                      
  \midrule                                                                                                                                                 
  \multicolumn{28}{l}{\textbf{Closed-source}} \\
  \rowcolor[HTML]{FBF6FB}
  GPT-5-Mini & \textbf{70.9} & \textbf{0.1} & 70.6 & 71.1 & 71.1 & 70.5 & 70.8 & 71.3 & \textcolor{red}{70.2} & 70.9 & 70.7 & 71.0 & 70.7 & 70.8 &         
  \textcolor{teal}{71.6} & 70.4 & 70.7 & 71.3 & 71.1 & 70.7 & 71.2 & 71.2 & 71.5 & 71.2 & 70.7 & 70.3 & 71.3 \\                                            
  \rowcolor[HTML]{FBF6FB}                                                                                                                                  
  GPT-5-Nano & 65.4 & 0.4 & 65.0 & 65.1 & 65.6 & 65.2 & 66.6 & \textcolor{teal}{67.0} & 66.0 & 65.3 & 64.6 & 64.9 & \textcolor{red}{64.2} & 64.9 & 66.1 &  
  64.9 & 65.3 & 65.5 & 65.1 & 65.7 & 66.1 & 65.1 & 65.7 & 65.0 & 65.4 & 65.4 & 65.8 \\                                                                     
  \rowcolor[HTML]{FBF6FB}                                                                                                                                  
  Gemini-2.5-Flash-Lite & 60.5 & 0.6 & 59.4 & 60.1 & 60.8 & 61.0 & 61.1 & 60.5 & 61.6 & 60.3 & 59.6 & 60.5 & 59.6 & 61.6 & 61.5 & 60.8 & 59.7 & 59.4 & 59.8 & 
  60.4 & 61.0 & \textcolor{teal}{61.7} & 60.5 & 60.9 & 60.8 & \textcolor{red}{58.6} & 60.5 \\                                                              
  \midrule
  \multicolumn{28}{l}{\textbf{Open-source}} \\                                                                                                             
  \rowcolor[HTML]{F5F5FF}
  Qwen3-VL-32B & \textbf{65.4} & 0.9 & 65.7 & 64.8 & 65.1 & 65.4 & 66.0 & 65.9 & 66.1 & 65.5 & 65.7 & 65.3 & 65.7 & 66.2 & 66.2 & \textcolor{teal}{66.3} & 
  \textcolor{red}{61.2} & 65.2 & 64.8 & 65.5 & 66.0 & 65.6 & 65.3 & 65.8 & 65.1 & 65.7 & 65.3 \\                                                           
  \rowcolor[HTML]{F5F5FF}
  Qwen3-VL-8B & 60.9 & 6.3 & 61.8 & 56.4 & 62.6 & 59.0 & 63.3 & 63.2 & 60.7 & 61.9 & 58.5 & 61.0 & 57.5 & \textcolor{teal}{63.4} & 61.8 & 62.5 &           
  \textcolor{red}{52.1} & 61.9 & 61.1 & 62.0 & 62.0 & 62.2 & 61.9 & 62.2 & 61.3 & 59.5 & 62.7 \\                                                           
  \rowcolor[HTML]{F5F5FF}
  Qwen3-VL-4B & 57.4 & 9.3 & 60.7 & 55.0 & 61.0 & 53.4 & 58.7 & 61.2 & 54.5 & 58.9 & 53.1 & 58.6 & 55.7 & \textcolor{teal}{61.8} & 59.4 & 60.9 &           
  \textcolor{red}{48.5} & 56.5 & 57.1 & 56.0 & 56.0 & 56.3 & 58.5 & 57.2 & 57.3 & 57.3 & 61.0 \\                                                           
  \rowcolor[HTML]{F5F5FF}                                                                                                                                  
  Qwen3-VL-30B-A3B & 61.1 & 5.7 & 62.2 & 56.5 & 60.2 & 60.6 & 62.6 & \textcolor{teal}{63.2} & 63.0 & 62.1 & 61.3 & 61.3 & 57.7 & 61.9 & 62.4 & 62.2 &      
  \textcolor{red}{52.0} & 61.2 & 62.3 & 61.2 & 61.5 & 62.6 & 62.5 & 62.1 & 61.0 & 61.9 & 62.8 \\                                                           
  \rowcolor[HTML]{F5F5FF}
  Qwen3-VL-2B & 51.4 & 3.9 & 51.3 & 48.6 & \textcolor{teal}{54.3} & 50.5 & 51.8 & 53.8 & 53.3 & 53.1 & 47.4 & 49.8 & 49.6 & 53.1 & 52.5 & 51.2 &           
  \textcolor{red}{46.3} & 51.9 & 51.0 & 51.5 & 53.2 & 49.5 & 51.5 & 54.1 & 51.6 & 50.9 & 52.4 \\                                                           
  \rowcolor[HTML]{F5F5FF}                                                                                                                                                           
  \rowcolor[HTML]{F5F5FF}
  InternVL-3.5-14B & 58.4 & 5.9 & 58.9 & 53.6 & \textcolor{teal}{62.4} & 56.6 & 57.9 & 61.7 & 60.7 & 59.9 & 55.8 & 58.2 & 58.0 & 59.6 & 59.9 & 61.1 &      
  \textcolor{red}{51.0} & 59.5 & 58.4 & 57.2 & 58.0 & 57.0 & 60.5 & 58.1 & 57.7 & 58.3 & 60.4 \\                                                           
  \rowcolor[HTML]{F5F5FF}
  InternVL-3.5-8B & 58.2 & 12.7 & 60.2 & 53.4 & 62.3 & 56.0 & 57.3 & \textcolor{teal}{64.3} & 56.6 & 59.5 & 49.7 & 57.4 & 56.7 & 61.5 & 61.1 & 61.0 &      
  \textcolor{red}{49.5} & 60.0 & 56.3 & 60.5 & 58.5 & 57.9 & 61.7 & 57.1 & 58.1 & 55.3 & 61.9 \\                                                           
  \rowcolor[HTML]{F5F5FF}
  InternVL-3.5-4B & 54.4 & 12.7 & 57.2 & 47.8 & \textcolor{teal}{60.1} & 50.9 & 55.2 & 59.4 & 55.3 & 56.2 & 47.3 & 54.6 & 53.5 & 58.4 & 55.1 & 56.8 &      
  \textcolor{red}{46.4} & 55.6 & 53.2 & 51.4 & 55.7 & 52.7 & 56.5 & 56.0 & 57.0 & 50.7 & 58.0 \\                                                           
  \rowcolor[HTML]{F5F5FF}
  InternVL-3.5-2B & 51.2 & 5.4 & 51.6 & 49.9 & \textcolor{teal}{57.0} & 49.4 & 53.8 & 54.9 & 51.1 & 52.7 & \textcolor{red}{44.9} & 52.1 & 51.3 & 53.1 &    
  52.1 & 50.5 & 48.2 & 48.7 & 49.0 & 51.0 & 51.4 & 50.4 & 53.5 & 51.5 & 50.5 & 49.5 & 51.3 \\                                                              
  \rowcolor[HTML]{F5F5FF}                                                                                                                                  
  InternVL-3.5-1B & 46.4 & 6.6 & 48.4 & 43.4 & \textcolor{teal}{51.1} & \textcolor{red}{42.0} & 43.3 & 47.4 & 46.6 & 44.2 & 43.8 & 49.5 & 48.3 & 48.5 &    
  49.5 & 47.7 & 45.0 & 48.2 & 46.9 & 42.6 & 47.1 & 45.4 & 43.4 & 47.2 & 48.2 & 43.0 & 50.2 \\                                                              
  \rowcolor[HTML]{F5F5FF}
  Gemma-3-27B & 61.9 & 0.9 & 62.5 & 61.3 & 62.2 & 62.2 & 63.0 & \textcolor{teal}{63.3} & 61.4 & 61.0 & 60.7 & 62.6 & 62.4 & 63.0 & 62.0 & 62.9 &           
  \textcolor{red}{59.3} & 62.7 & 62.2 & 62.0 & 61.7 & 60.2 & 61.7 & 61.3 & 61.6 & 61.8 & 62.8 \\                                                           
  \rowcolor[HTML]{F5F5FF}                                                                                                                                  
  Gemma-3-12B & 57.6 & 3.6 & 58.4 & 56.5 & 59.6 & 57.7 & 57.7 & 58.8 & 56.7 & 57.5 & 55.0 & 58.5 & 57.4 & \textcolor{teal}{60.2} & 59.0 & 58.6 &           
  \textcolor{red}{50.1} & 58.6 & 57.7 & 59.1 & 57.1 & 57.0 & 58.9 & 58.4 & 56.4 & 56.1 & 57.9 \\                                                           
  \rowcolor[HTML]{F5F5FF}
  Gemma-3-4B & 48.3 & 2.3 & 50.0 & 47.2 & 48.1 & 46.4 & 47.2 & 48.0 & 48.9 & 48.3 & 46.5 & 49.9 & \textcolor{teal}{52.1} & 51.0 & 47.0 & 49.3 &            
  \textcolor{red}{45.0} & 49.4 & 48.5 & 48.8 & 47.1 & 46.9 & 49.5 & 48.2 & 47.6 & 48.2 & 49.5 \\                                                           
  \rowcolor[HTML]{F5F5FF}                                                                                                                                  
  Pixtral-12B & 48.1 & 2.7 & 48.5 & 48.5 & 50.1 & 47.9 & 45.4 & 48.7 & 48.2 & 49.4 & 47.0 & 48.6 & 46.0 & 50.1 & 48.4 & 48.1 & 45.5 & 49.8 & 48.7 & 47.6 & 
  50.3 & \textcolor{red}{45.1} & 49.8 & \textcolor{teal}{51.3} & 45.7 & 46.6 & 48.3 \\                                                                     
  \bottomrule
  \end{tabular}                                                                                                                                            
  \vspace{-2mm}   
  \caption{\small{Overall performance on the M-MM-RewardBench subset. `Avg' and `Var' are computed across
  languages; lower is better for `Var'. \textcolor{teal}{Green} denotes the best and \textcolor{red}{Red} denotes the worst in each row. }}
  \label{tab:mvlrb_original_results}                                         
  \end{table*}   

 \vspace{-1mm}
\subsection{Additional Results}
\label{sec:mmrewardbench-benchmark}
 \vspace{-1mm}
Beyond using M-MM-RewardBench as a training resource, we benchmark LVLM judges on it to verify that the training corpus is substantive. Given its scale (100K samples), we exclude some proprietary models (GPT-5, Gemini-2.5-Flash, Claude-4.5-Haiku, Grok-4.1-Fast) due to high API cost, and report results for the remaining LVLMs in Table~\ref{tab:mvlrb_original_results}. We observe that model rankings largely mirror M-VL-RewardBench and M-OpenCQA: GPT-5-Mini leads overall (70.9\%), Qwen3-VL-32B tops among the open models (65.4\%) with consistent scaling, and Kazakh again yields the worst per-model accuracy for most LVLMs. One notable shift is that Gemma-3-27B becomes competitive (61.9\%), suggesting its weakness on M-VL-RewardBench is task-specific. This analysis provides additional evidence that our cross-lingual findings reflect genuine judge behavior. 
 \vspace{-1mm}
    \section{Conclusion and Future Work}

We introduce \textbf{MM-JudgeBench}, a benchmark for evaluating LVLM-as-a-judge models in multilingual and multimodal settings. It comprises two evaluation subsets, M-VL-RewardBench (general multimodal reasoning) and M-OpenCQA (chart-specific visual-text reasoning), spanning 25 languages with over 60K samples, complemented by a 100K-sample multilingual training set derived from MM-RewardBench.
Our evaluation of a broad range of closed and open models yields three key insights: (i) flagship closed models achieve the strongest overall accuracy and cross-lingual stability, though some of their optimized variants often degrade substantially; (ii) among open models, the Qwen3-VL family shows the most consistent scaling behavior, at times rivaling closed-source alternatives; and (iii) performance varies markedly across languages, with low-resource languages such as Kazakh posing persistent challenges. Our robustness analyses further reveal biases that are highly model- and dataset-dependent, underscoring that accuracy alone is insufficient to ensure reliable judging.
Overall, MM-JudgeBench exposes cross-lingual limitations of LVLM judges that remain invisible under English-only evaluation. While supervised fine-tuning on our training set improves open-model performance, future work will explore additional strategies such as reinforcement learning \cite{kaelbling1996reinforcement}, and extend MM-JudgeBench to more languages and evaluation tasks \cite{laskar2025improving}.
    
\section*{Limitations}

MM-JudgeBench is derived by translating English-centric datasets into 24 additional languages. Despite rigorous translation model selection and quality filtering, some linguistic or cultural nuances may be lost. However, by controlling translation quality with human intervention and keeping visual content fixed, MM-JudgeBench isolates cross-lingual judge behavior, enabling a focused study of multilingual generalization. Although the benchmark spans 25 typologically diverse languages, it does not cover many other low-resource languages. However, we address a critical gap in prior work by incorporating low-resource languages like Kazakh while demonstrating the limitations of most models in this language. Furthermore, our benchmark is primarily constructed with LLM-generated translations. Nonetheless, we conduct human\footnote{The human annotators possess strong English proficiency and hold graduate-level degrees from institutions where English is a primary academic language. } evaluation of the translated data (see Appendix \ref{human_eval_translation}) and find that humans also rate our multilingual datasets with high ratings. While OpenCQA preference labels are generated using GPT-5 and our human evaluation confirms agreement with it as a high-quality reference judge, there are still some risks of having reference-model bias. Nonetheless, we further validated the choice of GPT-5 by comparing cross-judge consistency with another stronger LVLM, Gemini-2.5-Pro (Spearman $\rho = 0.93$, Pearson $r = 0.94$). 

\section*{Acknowledgments}
We thank all the anonymous reviewers, the area chair, and the senior area chair of ACL 2026 for their excellent review comments. 
This research is supported by the Natural Sciences and Engineering Research Council (NSERC) of Canada, the York Research Chairs (YRC) program,  Canada Foundation for Innovation (CFI), Google's Gemini Academic Program for the API Credits, CUPE 3903 Research Grant, and Digital Research Alliance of Canada for the computing resources.

\section*{Ethics Statement}

This work evaluates automated vision–language judges rather than deploying them. Our findings reveal substantial cross-lingual variance and bias, highlighting risks in using LVLM judges as substitutes for human evaluation in multilingual settings.
MM-JudgeBench is designed to expose, not obscure, these risks by making multilingual failure modes measurable and transparent. Practitioners should avoid deploying judge models in high-stakes or user-facing scenarios without multilingual validation and human oversight.
All data are derived from publicly available benchmarks, contain no personally identifiable information, and are processed using automated translation. The licensing requirements are maintained accordingly while using different tools. Additional human compensation is not required since it was done by two authors of this paper. We also provide necessary details related to our human evaluation studies (see Appendix \ref{human_eval_opencqa} and \ref{human_eval_translation}). Finally, we used AI-based writing assistants only to improve the presentation of the paper.

\bibliography{custom}
\input{appendix}

\end{document}

%% file: vlrewardbench_results_table.tex
\begin{table*}[t]
\centering
\tiny
\setlength{\tabcolsep}{2pt}
\renewcommand{\arraystretch}{1}
\begin{tabular}{lccccccccccccccccccccccccccc}
\toprule
\rowcolor[HTML]{ECEFF1}
\textbf{Model} & \textbf{Avg} & \textbf{Var} & \textit{ar} & \textit{bn} & \textit{zh} & \textit{cs} & \textit{nl} & \textit{en} & \textit{fr} & \textit{de} & \textit{el} & \textit{he} & \textit{hi} & \textit{id} & \textit{it} & \textit{ja} & \textit{kk} & \textit{ko} & \textit{fa} & \textit{pl} & \textit{pt} & \textit{ro} & \textit{ru} & \textit{es} & \textit{tr} & \textit{uk} & \textit{vi} \\
\midrule
\rowcolor[HTML]{FAFAFA} \multicolumn{28}{l}
{\textbf{Closed-source}} \\ 
\rowcolor[HTML]{FBF6FB} 
GPT-5 & \textbf{81.3} & \textbf{0.2} & 81.4 & \textcolor{red}{80.6} & 81.6 & 81.6 & \textcolor{teal}{82.5} & 81.7 & 81.6 & 81.4 & 81.4 & 81.7 & 81.0 & 81.0 & 81.6 & 81.5 & \textcolor{red}{80.6} & 81.1 & 81.3 & 81.9 & 82.0 & 81.6 & \textcolor{red}{80.6} & 80.7 & 80.9 & 81.4 & 80.8 \\
\rowcolor[HTML]{FBF6FB} 
GPT-5-Mini & 78.1 & 0.4 & 78.4 & 76.8 & 77.8 & 78.1 & 78.6 & 78.6 & 78.1 & 78.4 & 78.5 & 78.2 & 77.6 & 77.4 & 78.7 & 78.6 & \textcolor{red}{76.6} & 78.2 & 77.5 & 78.6 & \textcolor{teal}{79.2} & 78.6 & 78.3 & 78.6 & 77.7 & 78.2 & 77.6 \\
\rowcolor[HTML]{FBF6FB} 
GPT-5-Nano & 73.2 & 1.2 & 73.5 & 72.1 & 73.9 & 73.5 & 74.3 & 71.2 & 74.3 & \textcolor{teal}{74.7} & \textcolor{red}{70.5} & 72.5 & 72.9 & 72.9 & 74.5 & 73.1 & 71.2 & 73.9 & 73.6 & 73.4 & 74.2 & 72.7 & \textcolor{teal}{74.7} & 73.8 & 73.4 & 73.3 & 72.6 \\
\rowcolor[HTML]{FBF6FB} Gemini-2.5-Flash & 76.7 & 0.5 & 77.6 & 75.4 & 77.0 & 77.2 & 76.7 & 78.6 & 76.0 & 76.4 & 76.4 & 77.3 & 76.7 & 77.3 & 77.6 & 76.0 & 75.7 & 77.0 & 75.9 & 77.1 & 76.7 & 76.7 & 77.1 & 76.8 & 76.3 & 76.0 & 76.9 \\
\rowcolor[HTML]{FBF6FB} 
Gemini-2.5-Flash-Lite & 40.8 & 2.6 & \textcolor{red}{35.8} & 39.5 & 40.3 & 42.0 & 41.8 & 40.2 & 41.7 & 40.3 & 39.7 & 40.2 & 41.7 & \textcolor{teal}{42.8} & 40.4 & 42.4 & 37.9 & 42.0 & 41.5 & 41.3 & 42.1 & 42.5 & 42.2 & 41.9 & 38.9 & 40.6 & 39.7 \\
\rowcolor[HTML]{FBF6FB} 
Claude-4.5-haiku & 56.4 & 2.1 & 57.5 & 55.7 & 55.3 & 55.9 & 55.9 & 54.6 & 58.3 & 57.1 & \textcolor{teal}{58.7} & 57.3 & 56.7 & 56.3 & 57.1 & 58.6 & \textcolor{red}{51.9} & 57.5 & 55.3 & 55.2 & 57.1 & 55.8 & 58.0 & 57.1 & 55.6 & 55.8 & 56.8 \\
\rowcolor[HTML]{FBF6FB} 
Grok-4.1-Fast & 71.3 & 0.7 & 72.4 & 70.4 & 70.2 & 71.7 & 72.7 & 72.0 & 71.9 & 71.5 & 70.8 & 71.3 & 71.3 & 71.0 & \textcolor{teal}{73.0} & 70.6 & \textcolor{red}{69.2} & 70.8 & 71.4 & 72.0 & 71.5 & 71.5 & 71.4 & 71.3 & 70.7 & 69.8 & 71.0 \\

\midrule

\rowcolor[HTML]{FAFAFA} \multicolumn{28}{l}{\textbf{Open-source}} \\ 
\rowcolor[HTML]{F5F5FF} 
Qwen3-VL-32B & \textbf{68.8} & 3.3 & 68.1 & 66.4 & 69.4 & 69.4 & 70.1 & 69.6 & 69.8 & 70.2 & 68.8 & 67.0 & 68.2 & 69.9 & \textcolor{teal}{70.5} & 69.1 & \textcolor{red}{62.0} & 68.8 & 68.1 & 69.8 & 70.1 & 69.6 & 70.2 & 69.9 & 66.9 & 69.8 & 69.2 \\
\rowcolor[HTML]{F5F5FF}
Qwen3-VL-8B & 62.2 & 1.8 & 61.8 & 59.8 & 62.6 & 62.8 & 63.1 & 62.9 & 62.9 & 62.9 & 60.3 & 61.2 & 61.8 & 61.9 & 63.1 & 63.4 & \textcolor{red}{58.5} & 61.6 & 62.2 & 62.1 & \textcolor{teal}{64.8} & 61.6 & 63.1 & 64.2 & 63.2 & 62.1 & 61.9 \\
\rowcolor[HTML]{F5F5FF}
Qwen3-VL-4B & 61.9 & 2.5 & 62.3 & 61.0 & 63.4 & 60.6 & 63.3 & 62.5 & \textcolor{teal}{64.4} & 62.8 & 60.9 & 59.9 & 61.2 & 61.2 & 63.1 & 63.6 & \textcolor{red}{57.3} & 63.4 & 60.6 & 60.5 & 64.3 & 61.2 & 62.3 & 63.0 & 61.7 & 60.9 & 62.7 \\
\rowcolor[HTML]{F5F5FF}
Qwen3-VL-30B-A3B & 63.4 & 2.2 & 63.5 & 61.3 & 63.7 & 62.6 & 64.4 & 65.3 & 64.7 & 63.9 & 62.2 & 63.3 & 62.2 & 63.8 & 65.1 & 63.8 & \textcolor{red}{58.3} & 63.7 & 62.6 & 63.8 & 64.9 & 62.8 & 64.0 & \textcolor{teal}{65.4} & 62.6 & 63.7 & 64.4 \\
\rowcolor[HTML]{F5F5FF}
Qwen3-VL-2B & 54.3 & 1.6 & 53.8 & \textcolor{teal}{55.7} & 53.1 & 53.7 & 54.8 & 55.5 & 55.2 & 55.3 & 53.6 & 52.2 & 55.4 & 55.4 & 54.9 & 54.5 & \textcolor{red}{50.5} & 53.4 & 52.7 & 53.8 & \textcolor{teal}{55.7} & 55.0 & 55.4 & 55.0 & 54.8 & 54.0 & 54.7 \\
\rowcolor[HTML]{F5F5FF}
InternVL-3.5-14B & 58.8 & 3.1 & 58.3 & 57.7 & 58.7 & 59.3 & 58.9 & \textcolor{teal}{62.0} & 59.9 & 59.9 & 57.5 & 59.0 & 57.8 & 57.7 & 59.3 & 57.8 & \textcolor{red}{53.3} & 58.5 & 59.6 & 58.5 & 61.4 & 59.6 & 60.2 & 60.9 & 55.7 & 58.7 & 59.1 \\
\rowcolor[HTML]{F5F5FF}
InternVL-3.5-8B & 52.9 & \textbf{0.9} & 53.7 & 53.8 & 51.8 & \textcolor{red}{51.6} & 52.6 & 53.8 & 52.1 & 51.9 & 54.6 & 53.3 & \textcolor{teal}{55.0} & 52.1 & 52.1 & 53.6 & 52.6 & 53.0 & 54.6 & 52.9 & 52.3 & 52.4 & 53.2 & 53.0 & 52.2 & 52.1 & 53.1 \\
\rowcolor[HTML]{F5F5FF}
InternVL-3.5-4B & 54.8 & 1.0 & 55.4 & 53.6 & 55.3 & 54.2 & 54.2 & \textcolor{teal}{57.1} & 54.8 & 55.2 & 56.9 & 54.2 & 55.5 & 54.0 & 54.4 & \textcolor{red}{53.0} & 54.2 & 54.7 & 54.8 & 53.7 & 56.0 & 55.0 & 55.2 & 55.9 & 53.7 & 54.2 & 53.8 \\
\rowcolor[HTML]{F5F5FF}
InternVL-3.5-2B & 53.6 & 1.2 & 53.5 & 52.6 & 52.8 & 54.1 & 53.9 & \textcolor{teal}{57.5} & 54.4 & 53.8 & 52.4 & 52.6 & 52.8 & 53.2 & 53.8 & 53.0 & 54.7 & 53.5 & \textcolor{red}{51.4} & 53.6 & 54.3 & 53.2 & 54.3 & 54.4 & 53.0 & 52.8 & 53.8 \\
\rowcolor[HTML]{F5F5FF}
InternVL-3.5-1B & 50.0 & 3.8 & 50.1 & 49.8 & 50.2 & 49.6 & 49.6 & 54.3 & 47.7 & 49.0 & \textcolor{teal}{54.9} & 50.1 & 49.8 & 47.7 & 49.1 & 49.6 & 54.3 & 50.2 & 51.3 & 48.8 & 50.6 & 48.4 & 49.3 & 49.5 & 50.4 & 48.8 & \textcolor{red}{47.0} \\
\rowcolor[HTML]{F5F5FF}
Gemma-3-27B & 45.0 & 4.0 & 43.7 & 42.4 & 45.0 & 47.4 & 46.0 & \textcolor{teal}{50.2} & 46.8 & 45.0 & 44.8 & 44.3 & 42.5 & 45.6 & 46.2 & 44.6 & \textcolor{red}{40.1} & 43.7 & 42.6 & 46.2 & 46.2 & 45.5 & 46.2 & 46.7 & 43.3 & 45.6 & 44.6 \\
\rowcolor[HTML]{F5F5FF}
Gemma-3-12B & 42.0 & 0.8 & 41.1 & 41.4 & 41.1 & 41.4 & 41.8 & 44.0 & 41.3 & 42.6 & \textcolor{teal}{44.4} & 42.3 & \textcolor{red}{40.7} & 41.7 & 41.6 & 41.4 & 41.5 & 42.2 & 42.9 & 41.8 & 42.5 & 43.1 & 42.3 & 41.9 & 41.4 & 42.5 & 41.3 \\
\rowcolor[HTML]{F5F5FF}
Gemma-3-4B & 34.9 & 1.7 & 35.9 & \textcolor{red}{31.7} & 33.9 & 35.4 & 34.6 & 36.1 & 35.3 & 35.7 & 36.3 & 33.8 & 31.8 & 35.9 & 34.6 & 34.8 & 34.0 & 33.5 & 35.5 & 35.3 & 35.5 & 35.9 & 33.6 & 36.1 & 35.6 & 33.8 & \textcolor{teal}{36.7} \\
\rowcolor[HTML]{F5F5FF}
LLaVA-Critic-7B & 49.5 & 3.4 & 49.1 & 49.8 & 50.3 & 51.6 & 48.6 & 50.1 & 47.4 & 49.2 & 51.6 & 49.2 & 51.8 & 48.5 & 48.8 & 48.8 & \textcolor{teal}{52.5} & 48.2 & 51.0 & 49.0 & 47.7 & 52.2 & 48.2 & \textcolor{red}{44.1} & 51.4 & 49.2 & 49.2 \\
\rowcolor[HTML]{F5F5FF}
Pixtral-12B & 41.7 & 1.5 & 39.7 & 41.6 & 43.3 & 42.6 & 41.7 & 43.2 & 43.1 & 42.7 & 42.4 & 40.7 & 41.6 & 41.4 & \textcolor{teal}{44.3} & 43.1 & 40.7 & 39.9 & 41.3 & 41.0 & 41.5 & 40.6 & 41.4 & 41.5 & 42.5 & \textcolor{red}{39.4} & 42.3 \\
\bottomrule
\end{tabular}
\vspace{-2mm}
\caption{\small{Overall Performance on the M-VL-RewardBench subset. Here, `Avg' and `Var' are computed across languages, and lower is better for `Var'.  \textcolor{teal}{Green} denotes the best and \textcolor{red}{Red} denotes the worst in each row.} \vspace{-2mm}}
\label{tab:vlrewardbench_results}
\end{table*}

%% file: opencqa_results_table.tex
\begin{table*}[t]
\centering
\tiny
\setlength{\tabcolsep}{2pt}
\renewcommand{\arraystretch}{1}
\begin{tabular}{lccccccccccccccccccccccccccc}
\toprule
\rowcolor[HTML]{ECEFF1}
\textbf{Model} & \textbf{Avg} & \textbf{Var} & \textit{ar} & \textit{bn} & \textit{zh} & \textit{cs} & \textit{nl} & \textit{en} & \textit{fr} & \textit{de} & \textit{el} & \textit{he} & \textit{hi} & \textit{id} & \textit{it} & \textit{ja} & \textit{kk} & \textit{ko} & \textit{fa} & \textit{pl} & \textit{pt} & \textit{ro} & \textit{ru} & \textit{es} & \textit{tr} & \textit{uk} & \textit{vi} \\
\midrule
\rowcolor[HTML]{F5F5FF}
Qwen3-VL-32B & \textbf{67.4} & 1.4 & 67.3 & 65.7 & 69.0 & 67.4 & 68.5 & \textcolor{teal}{70.9} & 67.6 & 67.1 & 66.9 & 67.0 & 67.0 & 67.8 & 68.4 & 66.7 & \textcolor{red}{64.7} & 66.1 & 66.7 & 67.2 & 68.0 & 67.8 & 67.7 & 68.3 & 66.2 & 67.0 & 67.8 \\
\rowcolor[HTML]{F5F5FF}
Qwen3-VL-8B & 64.5 & 0.5 & 65.2 & 65.2 & 64.6 & 64.8 & \textcolor{red}{63.3} & \textcolor{teal}{66.3} & 64.0 & 64.6 & 65.3 & 64.0 & 64.0 & 63.5 & 64.2 & 64.4 & 63.9 & 64.2 & 64.4 & 64.8 & 64.2 & 63.9 & 64.4 & 64.0 & 65.2 & 65.4 & 64.2 \\
\rowcolor[HTML]{F5F5FF}
Qwen3-VL-4B & 62.7 & 0.8 & 64.1 & 62.4 & 63.7 & 61.2 & 63.2 & \textcolor{teal}{64.2} & 62.9 & 62.3 & 62.4 & 63.0 & 61.8 & 63.6 & 62.6 & 63.3 & \textcolor{red}{60.3} & 62.9 & 61.8 & 62.4 & 63.3 & 63.1 & 63.0 & 63.0 & 62.0 & 62.6 & 62.7 \\
\rowcolor[HTML]{F5F5FF}
Qwen3-VL-30B-A3B & 63.7 & 0.9 & 64.1 & 62.4 & 64.7 & 64.6 & 63.8 & \textcolor{teal}{66.6} & 64.0 & 63.4 & 62.6 & 63.3 & 62.6 & 64.2 & 64.4 & 63.6 & \textcolor{red}{62.0} & 63.6 & 62.9 & 64.2 & 64.1 & 63.3 & 63.5 & 64.2 & 63.8 & 62.9 & 63.5 \\
\rowcolor[HTML]{F5F5FF}
Qwen3-VL-2B & 56.2 & 0.4 & 55.7 & \textcolor{teal}{57.7} & 56.0 & 55.7 & 56.0 & 57.0 & 56.1 & 56.5 & 55.7 & 55.2 & 57.4 & 56.5 & 56.6 & 56.5 & \textcolor{red}{55.2} & 56.7 & 55.7 & 56.2 & 55.6 & 56.4 & 55.6 & 56.9 & \textcolor{red}{55.2} & 56.2 & 56.3 \\
\rowcolor[HTML]{F5F5FF}
InternVL-3.5-14B & 66.3 & 2.1 & 65.9 & 64.1 & 67.8 & 66.7 & 65.8 & \textcolor{teal}{71.1} & 66.8 & 67.8 & 65.0 & 65.5 & 66.0 & 67.5 & 67.3 & 66.4 & \textcolor{red}{63.4} & 66.6 & 65.6 & 66.3 & 67.3 & 67.0 & 66.2 & 66.7 & 64.8 & 65.7 & 65.4 \\
\rowcolor[HTML]{F5F5FF}
InternVL-3.5-8B & 63.1 & 0.4 & 63.3 & 63.5 & 63.2 & 63.0 & 61.9 & \textcolor{teal}{64.6} & 63.7 & 63.1 & \textcolor{red}{61.7} & 62.3 & 63.6 & 62.8 & 64.2 & 63.3 & 62.2 & 63.5 & 62.3 & 63.0 & 63.2 & 62.6 & 63.0 & 63.7 & 63.2 & 63.4 & 63.3 \\
\rowcolor[HTML]{F5F5FF}
InternVL-3.5-4B & 62.7 & 2.0 & 63.1 & 62.2 & 63.3 & 61.1 & 64.7 & \textcolor{teal}{65.7} & 63.9 & 63.0 & 61.0 & 62.0 & 61.7 & 63.6 & 64.0 & 62.7 & \textcolor{red}{59.4} & 62.4 & 60.9 & 61.8 & 64.0 & 62.0 & 63.7 & 64.7 & 62.0 & 62.4 & 62.8 \\
\rowcolor[HTML]{F5F5FF}
InternVL-3.5-2B & 57.7 & 1.9 & 58.2 & \textcolor{red}{55.1} & 59.6 & 57.0 & 58.4 & \textcolor{teal}{60.6} & 59.6 & 58.0 & 56.4 & 57.3 & 56.0 & 57.2 & 58.2 & 59.3 & 56.0 & 58.4 & 55.5 & 56.5 & 58.6 & 58.9 & 58.0 & 58.0 & 57.7 & 56.8 & 57.4 \\
\rowcolor[HTML]{F5F5FF}
InternVL-3.5-1B & 55.3 & 2.4 & 55.0 & 53.2 & 56.7 & 54.3 & 55.4 & \textcolor{teal}{57.3} & 56.5 & 56.1 & 54.2 & 56.3 & 53.9 & 56.0 & 56.6 & 55.2 & \textcolor{red}{50.4} & 53.8 & 54.5 & 56.3 & 56.7 & 56.0 & 55.1 & \textcolor{teal}{57.3} & 54.0 & 56.5 & 55.4 \\
\rowcolor[HTML]{F5F5FF}
Gemma-3-27B & 57.0 & 3.5 & 58.3 & 55.3 & 57.8 & 57.4 & 56.5 & \textcolor{teal}{64.1} & 57.0 & 56.7 & 55.6 & 56.0 & 56.4 & 56.1 & 59.1 & \textcolor{red}{55.2} & \textcolor{red}{55.2} & 55.3 & 56.0 & 57.0 & 57.5 & 58.7 & 57.2 & 56.1 & 55.5 & 56.8 & 58.1 \\
\rowcolor[HTML]{F5F5FF}
Gemma-3-12B & 59.1 & \textbf{0.2} & 58.7 & 59.6 & 58.8 & 59.6 & 59.2 & \textcolor{teal}{60.4} & 59.3 & 59.3 & 59.0 & 59.7 & 58.8 & 59.0 & 58.7 & 58.5 & 58.8 & 58.5 & 58.9 & 59.7 & 58.7 & 59.0 & 58.8 & 59.8 & \textcolor{red}{58.4} & 58.6 & 58.7 \\
\rowcolor[HTML]{F5F5FF}
Gemma-3-4B & 57.0 & 0.6 & 57.1 & 55.7 & 57.4 & 57.7 & 57.4 & \textcolor{teal}{58.0} & 57.8 & 57.7 & 56.8 & 56.2 & 57.2 & 56.8 & 57.3 & 57.0 & 55.6 & 56.6 & 56.2 & 57.2 & 57.2 & 57.7 & 56.2 & 57.6 & 57.0 & \textcolor{red}{54.9} & 56.6 \\
\rowcolor[HTML]{F5F5FF}
LLaVA-Critic-7B & 57.0 & 2.3 & 57.6 & 56.1 & 56.9 & 56.6 & 57.2 & 56.0 & 58.8 & 57.6 & 55.4 & 56.4 & 56.4 & 57.9 & 58.2 & 54.2 & \textcolor{red}{53.9} & 59.1 & 56.6 & 57.6 & 57.2 & 58.2 & \textcolor{teal}{59.2} & \textcolor{teal}{59.2} & 54.3 & 57.2 & 58.4 \\
\rowcolor[HTML]{F5F5FF}
Pixtral-12B & 56.8 & 0.7 & 56.6 & 56.8 & 56.7 & 57.5 & 56.5 & \textcolor{teal}{60.0} & 57.0 & 57.3 & 57.4 & 56.4 & 56.1 & 56.4 & 56.8 & 57.2 & \textcolor{red}{55.9} & 56.4 & 56.3 & 57.0 & 57.7 & 57.2 & 56.4 & 56.9 & 56.0 & 56.0 & 56.5 \\
\bottomrule
\end{tabular}
\vspace{-2mm}
\caption{\small{Overall performance on the M-OpenCQA subset. `Avg' and `Var' are computed across languages, and lower is better for Var. \textcolor{teal}{Green} denotes the best and \textcolor{red}{Red} denotes the worst per row.} \vspace{-2mm}}
\label{tab:opencqa_results}
\end{table*}

%% file: appendix.tex
\appendix

\section{Appendix}

\subsection{Translation Related Details}
\label{translation_details}
We demonstrate the translation prompt below. 

\definecolor{attachedColor5}{HTML}{ECEFF1}
\definecolor{attachedColor}{HTML}{e0efff}
\definecolor{attachedColor2}{HTML}{f3f3f3}
\definecolor{attachedColor3}{HTML}{FFE5CC}
\definecolor{attachedColor4}{HTML}{FFCCCC}
\begin{tcolorbox}[
boxrule=0.25pt,   
  colback=attachedColor2,    
  colframe=black,           
  colbacktitle=attachedColor, 
  coltitle=black,           
  title={{Translation Prompt}},
  fonttitle=\bfseries,      
  fontupper=\small          
]

Translate the following texts  (Query, Response\_1, Response\_2) into the following languages: \\ \\
French, Portuguese, Greek, Chinese, Hindi, Arabic, Czech, German, Hebrew, Indonesian, Italian, Japanese, Korean, Dutch, Persian, Polish, Romanian, Russian, Spanish, Turkish, Ukrainian, Vietnamese, Bengali, and Kazakh. \\ \\
Provide your response in an array of JSON format with the following keys for each JSON object: (i) language, (ii) Translated\_Query, (iii) Translated\_Response\_1, (iv) Translated\_Response\_2

\end{tcolorbox}


An example of the output format of Gemini-3-Pro translation is demonstrated below.
\begin{tcolorbox}[
    colback=gray!5,                  
    colframe=black,                  
    coltitle=white,                  
    fonttitle=\bfseries\Large,
    title={JSON Formatted Output},
    arc=1mm,                         
    boxrule=1pt,
    left=5mm, right=5mm, top=5mm, bottom=5mm
    %
]
\ttfamily\footnotesize  
[\\
\ \ \{ \\
\ \ \ \ \textbf{"Language"}: "French",\\
\ \ \ \ \textbf{"Translated\_Query"}: "Quels sont les principaux objets ou sujets de l'image ? Veuillez les décrire en détail.",\\
\ \ \ \ \textbf{"Translated\_Response\_1"}: "Les principaux objets de l'image sont un chien avec une laisse orange attachée à un banc. Le cadre semble être en extérieur, peut-être dans un environnement semblable à un parc, car on voit de l'herbe et un banc en bois.",\\
\ \ \ \ \textbf{"Translated\_Response\_2"}: "Les principaux objets de l'image sont un chien, un banc en bois portant l'inscription « WATSON BOWL » et une table ou un banc de pique nique. Le chien est assis sur l'herbe à côté du banc et semble attaché ou retenu par une laisse orange fixée au banc."\\
\ \ \}, \\
\ \ . . . . . . . . . . . . . . . . . . . . \\
\\
\ \ . . . . . . . . . . . . . . . . . . . . \\
\ \ \{ \\
\ \ \ \ \textbf{"Language"}: "German",\\
\ \ \ \ \textbf{"Translated\_Query"}: "Die Hauptobjekte im Bild sind ein Hund mit einer orangefarbenen Leine, die an einer Bank befestigt ist. Die Umgebung scheint im Freien zu sein, möglicherweise in einer parkähnlichen Umgebung, da man Gras und eine Holzbank sehen kann.",\\
\ \ \ \ \textbf{"Translated\_Response\_1"}: "Die Hauptobjekte im Bild sind ein Hund mit einer orangefarbenen Leine, die an einer Bank befestigt ist. Die Umgebung scheint im Freien zu sein, möglicherweise in einer parkähnlichen Umgebung, da man Gras und eine Holzbank sehen kann."\\
\ \ \ \ \textbf{"Translated\_Response\_2"}: "Die Hauptobjekte im Bild sind ein Hund, eine Holzbank mit der Aufschrift „WATSON BOWL“ sowie ein Tisch oder eine Picknickbank. Der Hund sitzt auf dem Gras neben der Bank und scheint mit einer orangefarbenen Leine, die an der Bank befestigt ist, angebunden oder zurückgehalten zu werden."\\
\}, \\
]
\end{tcolorbox}

\subsection{Cross-Lingual Amplification of Positional Bias}
\label{cross_lingual_amplification_pos_bias}

We additionally compare positional bias in English versus non-English languages across model families in Table \ref{tab:cross_lingual_pos_bias} and observe that positional bias consistently increases in non-English settings across all three model families, with InternVL-3.5 and Gemma-3 showing nearly 2x amplification. 

\begin{table}[t]
\centering
\scriptsize
\setlength{\tabcolsep}{4pt}
\begin{tabular}{l|c|c}
\toprule
  \rowcolor[HTML]{ECEFF1} \textbf{Model Family} & \textbf{Avg.\ English Pos.\ Bias} & \textbf{Avg.\ Non-English Pos.\ Bias} \\
\midrule
\rowcolor[HTML]{FAFAFA}
Qwen3-VL     & 11.03 & 14.44 \\
\rowcolor[HTML]{FAFAFA} InternVL-3.5 & 15.79 & 31.18 \\
\rowcolor[HTML]{FAFAFA}
Gemma-3      & 5.11  & 9.95  \\
\bottomrule
\end{tabular}
\caption{\small{Comparison of average positional bias in English versus non-English languages across three model families.}}
\label{tab:cross_lingual_pos_bias}
\end{table}

\subsection{Length Bias Analysis}
\label{length-bias-appendix}

\begin{figure}
    \centering\includegraphics[width=\linewidth]{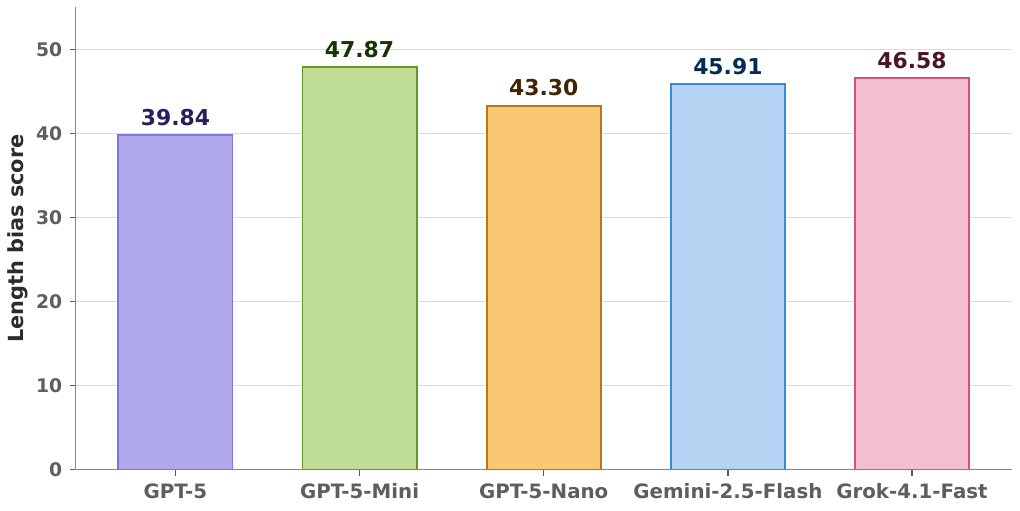}
    \caption{Length Bias in M-VL-RewardBench for closed-source models. Lower values indicate better.}
    \label{fig:length_bias}
\end{figure}

We show the length bias for the open models in Table \ref{tab:ap_len_bias} and for the closed models in Figure \ref{fig:length_bias}. For the open models, we find that the LVLMs tend to show higher length bias in MM-OpenCQA (above 50\%) than M-VL-RewardBench (below 50\%). Overall, we observe Qwen3-VL-2B demonstrates the lowest length bias. In terms of closed models, we find that the GPT-5 model shows the lowest length bias, while GPT-5-Mini shows the highest. 

\begin{table}[t!]
\tiny
\centering
\begin{tabular}{lccc}
\toprule
  \rowcolor[HTML]{ECEFF1} \textbf{Model} & \textbf{M-OpenCQA} & \textbf{M-VL-RewardBench} & \textbf{Average} \\
\midrule
Qwen3-VL-32B        & 65.38 & 39.25 & 52.32 \\
Qwen3-VL-8B         & 64.55 & 36.36 & 50.46 \\
Qwen3-VL-4B         & 63.82 & 40.15 & 51.99 \\
Qwen3-VL-30B-A3B    & 69.63 & 39.37 & \redval{54.50} \\
Qwen3-VL-2B         & 51.75 & \tealval{34.11} & \tealval{42.93} \\
InternVL-3.5-14B    & 66.16 & 37.39 & 51.78 \\
InternVL-3.5-8B     & 67.31 & 39.43 & 53.37 \\
InternVL-3.5-4B     & 61.58 & 36.66 & 49.12 \\
InternVL-3.5-2B     & 51.82 & 37.30 & 44.56 \\
InternVL-3.5-1B     & \tealval{50.34} & 37.15 & 43.25 \\
Gemma-3-27B         & \redval{70.77} & 37.14 & 53.96 \\
Gemma-3-12B         & 69.83 & 37.61 & 53.72 \\
Gemma-3-4B          & 63.72 & 39.70 & 51.71 \\
LLaVA-Critic-7B     & 58.56 & 35.08 & 46.82 \\
Pixtral-12B         & 65.13 & \redval{42.99} & 54.06 \\

\bottomrule
\end{tabular}
\caption{Average Length Bias for open-source LVLMs. Lower values indicate better with \textcolor{teal}{Green} indicates the best and \textcolor{red}{Red} indicates the worst, per column. }
\label{tab:ap_len_bias}
\end{table}

\subsection{Output Format Instruction Following Behavior Analysis.}
\label{output_format_instruction_appendix}
\begin{figure}[t!]
    \centering
    \includegraphics[width=\linewidth]{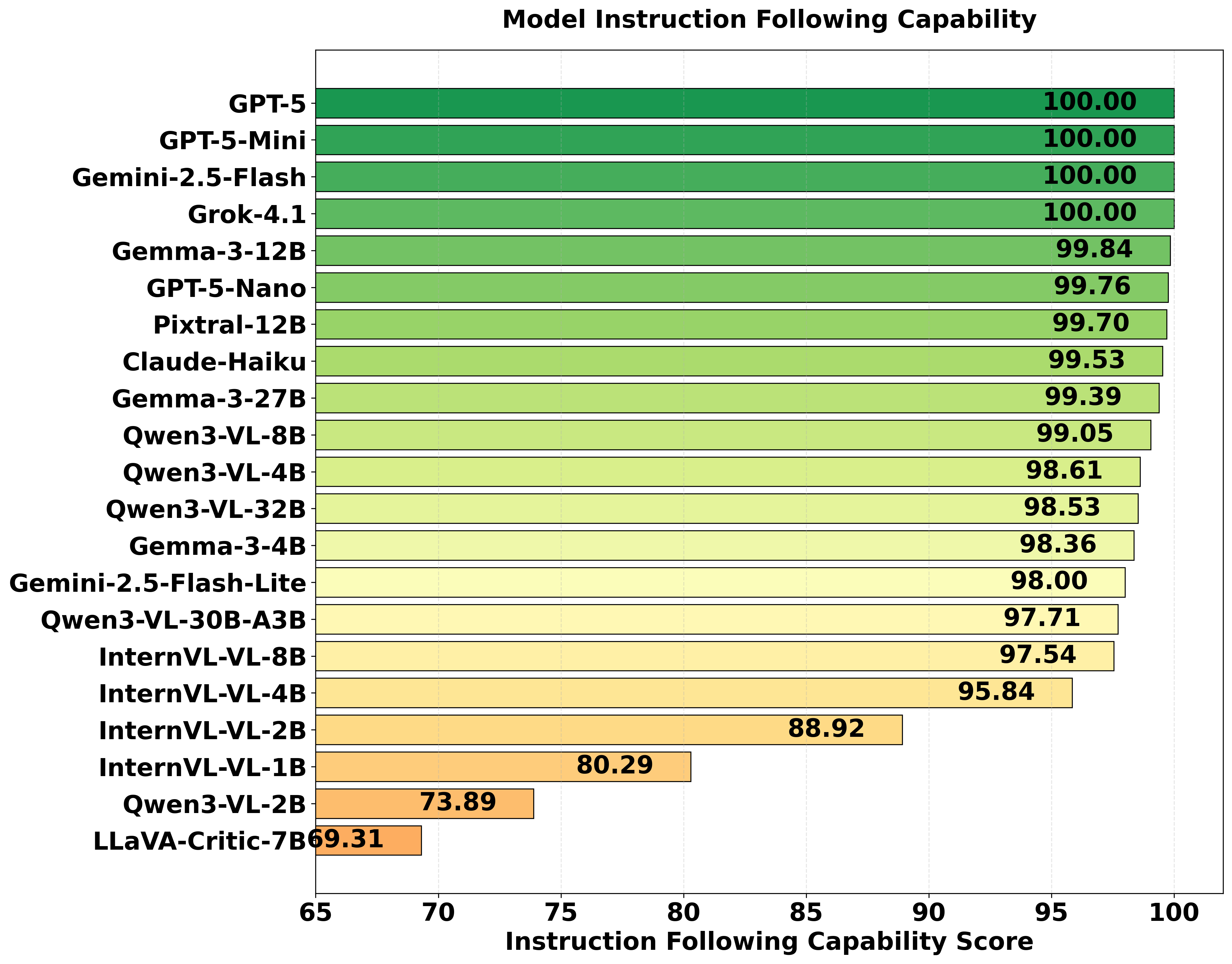}
    \caption{\small{Output Instruction Following Accuracy in M-VL-RewardBench.}}
    \label{fig:mm-reward-instruction}
\end{figure}

We evaluate instruction-following fidelity by measuring compliance with the required JSON output format in multilingual settings (Figure \ref{fig:mm-reward-instruction}). We find most closed models exhibit 100\% format following adherence, with Claude-4.5-Haiku and GPT-5-Nano achieving slightly below 100\% and the Gemini-2.5-Flash-Lite showing modest degradation (98\%). Among open models, while none of them could reach 100\% accuracy, almost all of them with more than 2B parameters achieve above 95\% accuracy (with Gemma-3-12-B achieving the best 99.84\% accuracy). Meanwhile, models below 2B parameters show poorer output format instruction following accuracy (below 90\% accuracy). Surprisingly, LLaVA-Critic-7B, despite being specifically trained for reward modeling, achieves only 69.3\% accuracy. This highlights that specialized training can make certain reward models weaker at following instructions. 

\subsection{Human Evaluation Instructions in OpenCQA}
\label{human_eval_opencqa}

Below, we show the instructions for our human evaluation in the OpenCQA dataset. 

\definecolor{attachedColor5}{HTML}{ECEFF1}
\definecolor{attachedColor}{HTML}{e0efff}
\definecolor{attachedColor2}{HTML}{f3f3f3}
\definecolor{attachedColor3}{HTML}{FFE5CC}
\definecolor{attachedColor4}{HTML}{FFCCCC}
\begin{tcolorbox}[
boxrule=0.25pt,   
  colback=attachedColor2,    
  colframe=black,           
  colbacktitle=attachedColor3, 
  coltitle=black,           
  title={{Instructions for Humans to Evaluate GPT-5 Judgments in M-OpenCQA}},
  fonttitle=\bfseries,      
  fontupper=\small          
]

You are given an image and a question related to the image. You are also provided with two candidate answers, A and B, generated by two models. \\

Your objective is to decide which answer is better based on the following criteria. \\

- Correctness with respect to the image and question.\\
- Completeness and level of detail.\\
- Relevance and clarity (no unnecessary verbosity).\\
- Avoiding hallucinations or unsupported claims.\\
\end{tcolorbox}

\subsection{Human Evaluation of the Translation Quality}
\label{human_eval_translation}

We conduct human evaluation on our MM-JudgeBench benchmark (on both subsets) to assess the quality of the translated data. However, given the difficulty in finding high-quality human evaluators having expertise across these 25 languages, we translated both the M-VL-RewardBench and M-OpenCQA datasets from the 24 non-English languages back to English. After back-translation, we check the BERTScore \cite{zhangbertscore} by comparing the original English data with the back-translated English data and observe 90\% similarity score.

Afterwards, we sample about 200 examples covering all languages from our benchmark and instruct the human annotator who has expertise in the English language to identify the quality of each example after back translation. We specifically ask the human annotator to evaluate based on the following criteria:

\begin{itemize}
    \item \textbf{Perfect:} The back-translated data is perfect without any translation errors.
    \item \textbf{Good:} The back-translated data does not have any major issues, but has some minor issues in terms of wording/synonyms, etc. However, this should not change the context of the original data. 
    \item \textbf{Bad:} The back-translated data has some major issues (e.g., factual errors) that affect the content of the original data.
\end{itemize}

Based on our human evaluation, we find that the human annotator rates \textbf{55.7\%} of the data as perfect, \textbf{43.8\%} as good, and only \textbf{0.5\%} of the samples as bad. This demonstrates the high quality of Gemini-3-Pro translation. 

To justify the reliability of the human evaluation via back translation, we have further conducted a native-speaker evaluation for Bengali across 300 samples. We find that the native speaker rated 73\% of the samples as Perfect and 27\% as Good, with no samples labeled as Bad. 

\subsection{Translation Quality of M-OpenCQA Based on LaBSE and CometKiwi}
\label{trans_quality_eval_opencqa}
We compute the LaBSE score and the CometKiwi score in the M-OpenCQA dataset and find that it has 0.86 LaBSE and 0.84 CometKiwi scores, which are well above our quality filtering threshold of 0.75. We also observe that the LaBSE and CometKiwi scores are above 0.8 across all languages in our M-OpenCQA dataset, without having any significant differences. Note that we also observe similar trend in M-VL-RewardBench.

\subsection{Example from M-OpenCQA}
\begin{figure*}[t!]
    \centering
    \includegraphics[width=\textwidth]{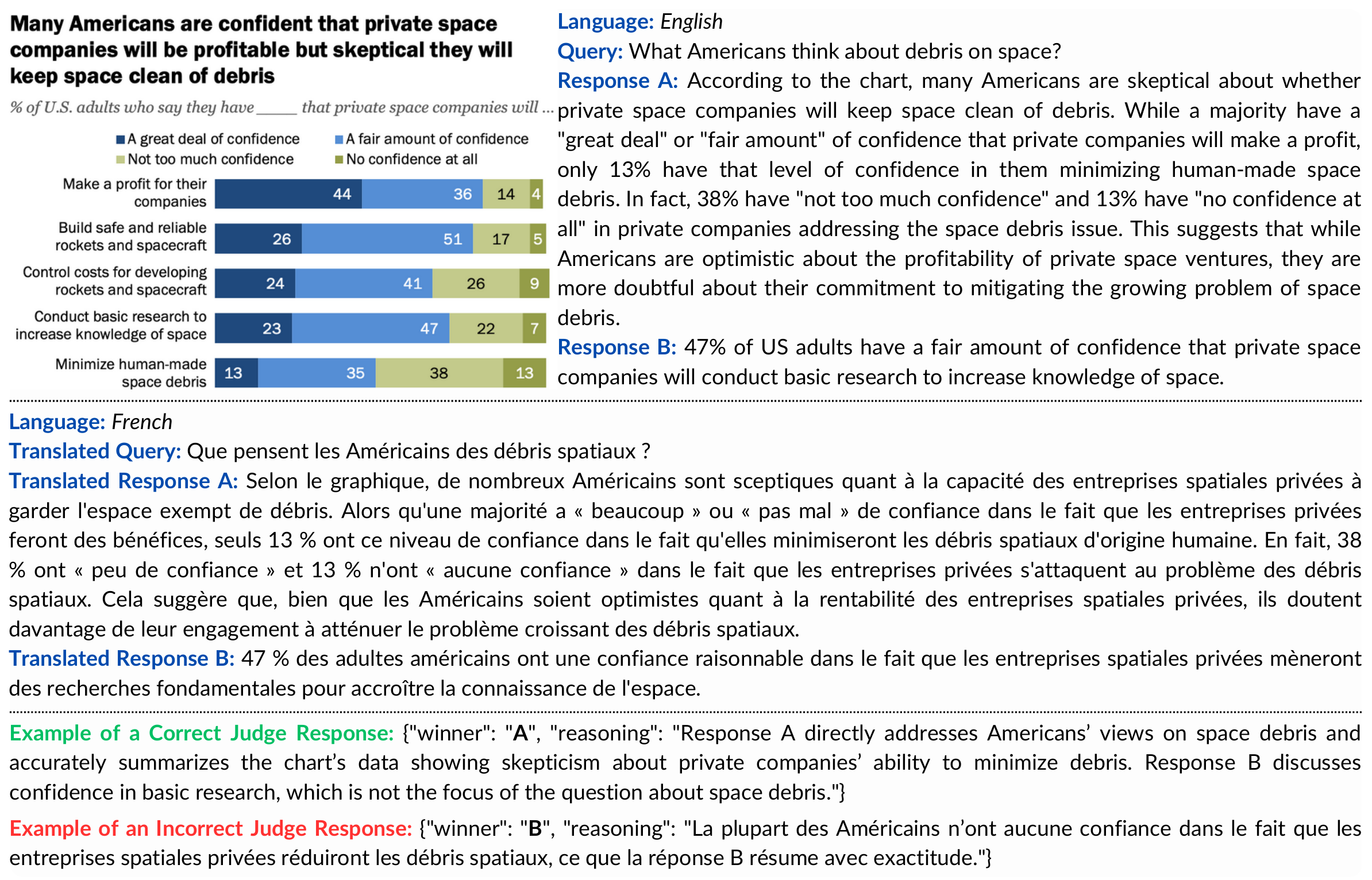}
    \caption{Example from MM-JudgeBench (OpenCQA) illustrating multilingual reward evaluation. A chart image–question pair with candidate responses is translated from English into French, where Response A accurately reflects the chart's content. The LVLM judge selects the correct response in English but incorrect in French, highlighting the need for language-invariant, visually grounded judgment.}
    \label{fig:opencqa_ex}
\end{figure*}

An example from M-OpenCQA in our MM-JudgeBench benchmark is shown in Figure \ref{fig:opencqa_ex}.

\subsection{English vs Non-English Performance Gap Across Tasks} 
\label{task_level_performance_gap}
To analyze multilingual robustness, we report the average accuracy gap between English and non-English languages, computed across comparable model sizes (2B/4B/8B for Qwen3-VL and InternVL-3.5; Mini/Nano for GPT) in Table \ref{tab_per_task_cross_lingual_degredation}. Positive values indicate higher performance in English, while negative values indicate better performance in non-English languages.

\begin{table}[t!]
\centering
\small
\setlength{\tabcolsep}{2pt}
\begin{tabular}{l|c|c|c}
\toprule
  \rowcolor[HTML]{ECEFF1} \textbf{Model Family} & \textbf{Hallucination} & \textbf{Mathmatical} &
\textbf{General} \\
\midrule
\rowcolor[HTML]{FAFAFA}
InternVL-3.5 & 2.95 & 1.19 & 1.39 \\
\rowcolor[HTML]{FAFAFA} Qwen3-VL    & 0.34 & 1.99 & 0.07 \\
\rowcolor[HTML]{FAFAFA}
GPT Series  & -1.18 & -0.24 & 3.53 \\
\bottomrule
\end{tabular}
\caption{Cross-lingual performance degradation per task in M-VL-RewardBench}
\label{tab_per_task_cross_lingual_degredation}
\end{table}

InternVL-3.5, which shows the largest gap on Hallucination Queries, while Qwen3-VL exhibits its largest gap on Mathematical Reasoning. Qwen3-VL maintains near-parity on General Multimodal tasks (0.07), whereas InternVL and GPT show larger degradation. Interestingly, GPT models slightly improve in non-English settings for Hallucination and Math, suggesting potential benefits from broader multilingual pretraining. Overall, multilingual consistency depends strongly on both task taxonomy and model family.

\subsection{Performance Comparison Based on Script Groups and Resources}
\label{performance_script_groups}
This section provides a performance comparison based on the seven language script groups\footnote{Each script group represents a distinct writing system with unique linguistic characteristics that could influence model performance.} and the resourcedness\footnote{High, Medium, and Low.}. We select the best open-source series model (Qwen3) and the best closed-source series model (GPT-5 series) and compare the performance gap between models in M-VL-RewardBench based on the corresponding efficiency tiers: GPT-5 vs Qwen3-32B, GPT-5-Mini vs Qwen3-8B, and GPT-5-Nano vs Qwen3-4B. Below, we first describe these script groups. 

\paragraph{Latin Script}
Latin script comprises 13 languages: English, French, German, Polish, Portuguese, Romanian, Spanish, Czech, Dutch, Turkish, Vietnamese, Indonesian, and Italian. It uses the Roman alphabet (A-Z) with diacritical marks and is the most widely used writing system globally. Latin script is left-to-right and phonetic, making it straightforward for tokenization.

\paragraph{Cyrillic Script}
Cyrillic script includes 3 languages: Russian, Ukrainian, and Kazakh. Derived from the Greek alphabet, it is used primarily in Eastern Europe and Central Asia. This group exhibits linguistic diversity, ranging from high-resource (Russian) to low-resource (Kazakh) languages. The script requires separate tokenization strategies from Latin.

\paragraph{Greek Script}
Greek script represents a single language: Greek. It is an ancient alphabet with unique character sets and diacritical marks. Greek has been extensively studied in linguistics and holds significant cultural importance.

\paragraph{Arabic Script}
Arabic script includes 2 languages: Arabic and Persian. It is a right-to-left writing system used primarily in the Middle East and parts of Asia. Arabic script includes diacritical marks and represents some of the most widely spoken languages in the world.

\paragraph{Hebrew Script}
Hebrew script represents a single language: Hebrew. It is a right-to-left bidirectional writing system with unique properties and historical significance. Hebrew includes diacritical marks (vowel points) and presents specific challenges for tokenization due to its bidirectional nature.

\paragraph{Devanagari Script}
Devanagari script includes 2 languages: Hindi and Bengali. 
These languages have significant speaker populations but relatively lower resource availability compared to Latin script languages.

\paragraph{CJK Scripts}
CJK scripts include 3 languages: Chinese, Japanese, and Korean. Chinese uses logographic characters, Japanese combines logographic kanji with syllabic scripts, and Korean uses the alphabetic hangul script. These languages present unique challenges for tokenization due to large character sets and complex morphology.

\begin{figure}
    \centering
    \includegraphics[width=\linewidth]{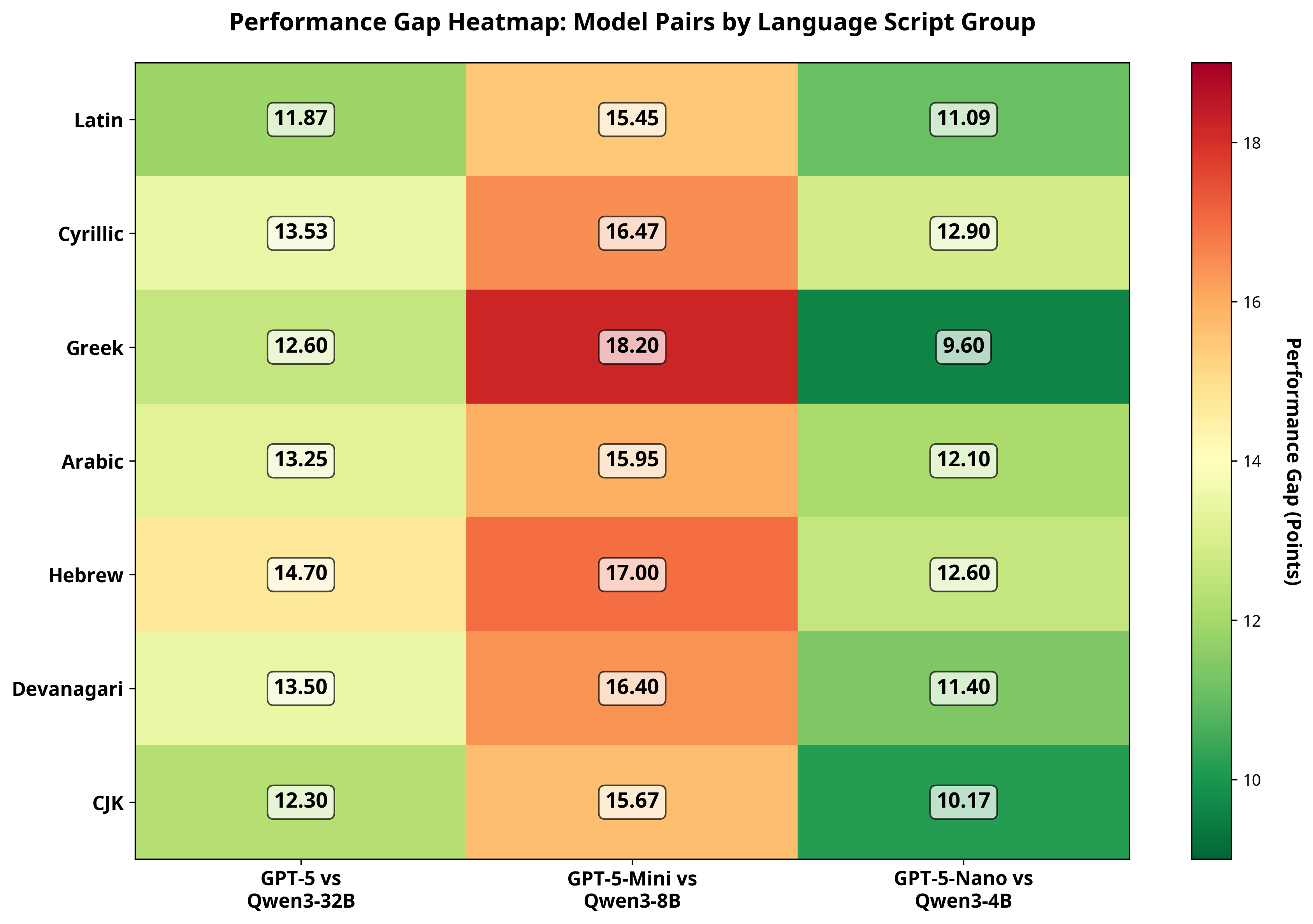}
    \caption{\small{Performance Comparison based on Scripts Groups.}}
    \label{fig:results_scripts_groups}
\end{figure}

\paragraph{} Figure \ref{fig:results_scripts_groups} shows a consistent closed–open gap across all script groups, but the gap is largest for the mid-tier pairing (GPT-5-Mini vs Qwen3-8B) across every script (between 15.45 to 18.20 points), suggesting the relative advantage of the closed model is most pronounced in that tier rather than monotonically increasing with ``model size.''

\paragraph{} In terms of resource availability, we categorize the languages following \texttt{Aya-101}\footnote{\url{https://huggingface.co/CohereLabs/aya-101}}, as demonstrated below:

\begin{itemize}
    \item 
\textbf{High-Resource (15 languages):}
English, Chinese, Japanese, Spanish, French, German, Italian, Portuguese, Russian, Dutch, Arabic, Polish, Hindi, Persian, Czech

  \item  \textbf{Mid-Resource (9 languages):}
Korean, Turkish, Vietnamese, Indonesian, Greek, Hebrew, Romanian, Ukrainian, Bengali

  \item  \textbf{Low-Resource (1 language):}
Kazakh
\end{itemize}

\begin{figure}
    \centering
    \includegraphics[width=\linewidth]{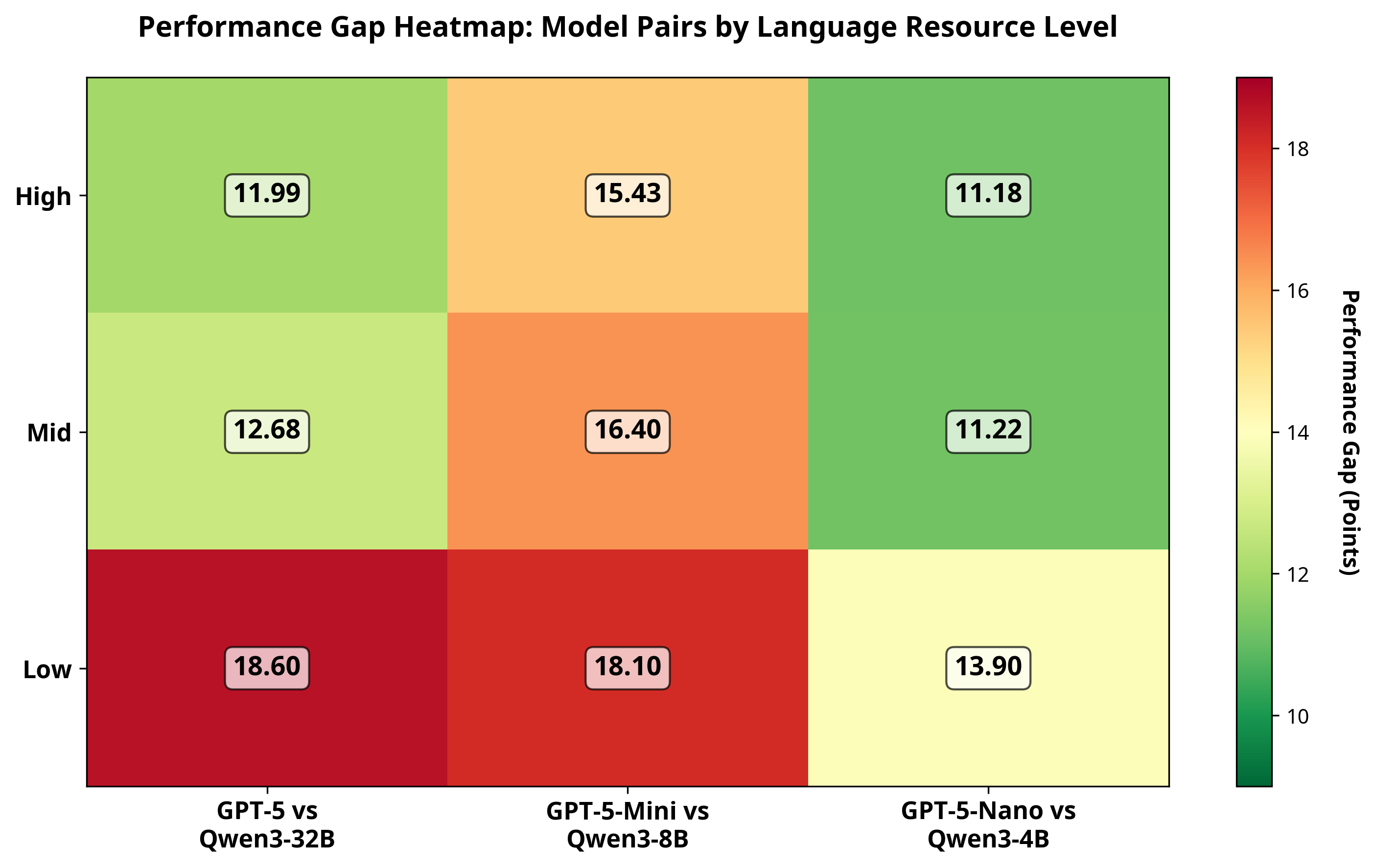}
    \caption{\small{Performance Comparison based on Resource Level.}}
    \label{fig:results_resource}
\end{figure}

Figure \ref{fig:results_resource} shows that performance gaps between GPT-5 and Qwen3 models grow as language resources decrease, with the largest gaps consistently appearing for low-resource languages. The mid-tier pairing (GPT-5-Mini vs Qwen3-8B) exhibits the strongest gap across all resource levels, while the nano vs Qwen3-4B tier shows the smallest and most stable differences, indicating closer competitiveness at smaller scales.